\documentclass[pdflatex,sn-nature]{sn-jnl}% referee option is meant for double line spacing
\usepackage{subfigure} 
\usepackage{graphicx}%
\usepackage{multirow}%
\usepackage{amsmath,amssymb,amsfonts}%
\usepackage{amsthm}%
\usepackage{mathrsfs}%
\usepackage[title]{appendix}%
\usepackage{xcolor}%
\usepackage{textcomp}%
\usepackage{manyfoot}%
\usepackage{booktabs}%
\usepackage{algorithm}%
\usepackage{algorithmicx}%
\usepackage{algpseudocode}%
\usepackage{listings}%
\usepackage[percent]{overpic} % for text over images
\usepackage{color}

\usepackage{subfiles}
\usepackage{cleveref}
\usepackage{xr-hyper}

\newcommand{\rr}{\color{red}}
\newcommand{\bb}{\color{black}}

\newcommand{\bl}{\color{black}}

\newcommand{\mycomment}[1]{}

\theoremstyle{thmstyleone}%
\theoremstyle{thmstyletwo}%

\theoremstyle{thmstylethree}%

\makeatletter

\newcommand*{\addFileDependency}[1]{% argument=file name and extension
  \typeout{(#1)}
  \@addtofilelist{#1}
  \IfFileExists{#1}{}{\typeout{No file #1.}}
}

\makeatother

\newcommand*{\myexternaldocument}[1]{%
    \externaldocument{#1}%
    \addFileDependency{#1.tex}%
    \addFileDependency{#1.aux}%
}

\myexternaldocument{supplementary}

\begin{document}

\title[Article Title]{Handling Missing Modalities in Multimodal Survival Prediction for Non-Small Cell Lung Cancer}

\author[1,2]{\fnm{Filippo} \sur{Ruffini}}\email{filippo.ruffini@unicampus.it}
\equalcont{F.R. and C.M.C. contributed equally to this work. S.R. and V.G. and P.S. jointly supervised this work.}

\author[1]{\fnm{Camillo} \sur{Maria Caruso}}
\equalcont{F.R. and C.M.C. contributed equally to this work. S.R. and V.G. and P.S. jointly supervised this work.}
\author[4]{\fnm{Claudia} \sur{Tacconi}}
% Fill here author info
\author[5, 6]{\fnm{Lorenzo} \sur{Nibid}}
\author[10]{\fnm{Francesca} \sur{Miccolis}}
\author[10]{\fnm{Marta} \sur{Lovino}}
\author[3, 4]{\fnm{Carlo} \sur{Greco}}
\author[3, 4]{\fnm{Edy} \sur{Ippolito}}
\author[3, 4]{\fnm{Michele} \sur{Fiore}}
\author[7,8]{\fnm{Alessio} \sur{Cortellini}}
\author[9]{\fnm{Bruno} \sur{Beomonte Zobel}}
\author[5, 6]{\fnm{Giuseppe} \sur{Perrone}}
\author[7, 8]{\fnm{Bruno} \sur{Vincenzi}}
\author[11]{\fnm{Claudio} \sur{Marrocco}}
\author[11]{\fnm{Alessandro} \sur{Bria}}
\author[10]{\fnm{Elisa} \sur{Ficarra}}
\author[3,4]{\fnm{Sara} \sur{Ramella}}
\equalcont{F.R. and C.M.C. contributed equally to this work. S.R. and V.G. and P.S. jointly supervised this work.}

\author[1]{\fnm{Valerio} \sur{Guarrasi}}\email{valerio.guarrasi@unicampus.it}
\equalcont{F.R. and C.M.C. contributed equally to this work. S.R., V.G. and P.S. jointly supervised this work.}

% Equal contribution Sara - Vale - Paolo
% Valerio qui?
\author[1,2]{\fnm{Paolo} \sur{Soda}}\email{paolo.soda@umu.se}
\equalcont{F.R. and C.M.C. contributed equally to this work. 
S.R., V.G. and P.S. jointly supervised this work.}

\affil[1]{
\orgdiv{Unit of Artificial Intelligence and Computer Systems, Department of Engineering}, 
\orgname{Università Campus Bio-Medico di Roma}, \orgaddress{\city{Rome}, \country{Italy}}}
\affil[2]{\orgdiv{Department of Diagnostics and Intervention, Radiation Physics, Biomedical Engineering}, \orgname{Umeå University}, \orgaddress{\city{Umeå}, \country{Sweden}}}

\affil[3]{\orgdiv{Research Unit of Radiation Oncology, Department of Medicine and Surgery}, \orgname{Università Campus Bio-Medico di Roma}, \orgaddress{\city{Rome}, \country{Italy}}}
\affil[4]{\orgdiv{Operative Research Unit of Radiation Oncology}, \orgname{Fondazione Policlinico Universitario Campus Bio-Medico}, \orgaddress{\city{Roma}, \country{Italy}}}

\affil[5]{\orgdiv{Anatomical Pathology Operative Research Unit}, \orgname{Fondazione Policlinico Universitario Campus Bio-Medico}, \orgaddress{\city{Rome}, \country{Italy}}}
\affil[6]{\orgdiv{Research Unit of Anatomical Pathology, Department of Medicine and Surgery}, \orgname{Università Campus Bio-Medico di Roma}, \orgaddress{\city{Rome}, \country{Italy}}}

\affil[7]{\orgdiv{Department of Medicine and Surgery}, \orgname{Università Campus Bio-Medico di Roma},\orgaddress{\city{Roma}, \country{Italy}}}
\affil[8]{\orgdiv{Operative Research Unit of Medical Oncology}, \orgname{Fondazione Policlinico Universitario Campus Bio-Medico}, \orgaddress{\city{Roma}, \country{Italy}}}

\affil[9]{\orgdiv{Operative Research Unit of Radiology and Interventional Radiology}, \orgname{Fondazione Policlinico Universitario Campus Bio-Medico}, \orgaddress{\city{Roma}, \country{Italy}}}

\affil[10]{\orgdiv{Department of Engineering ‘Enzo Ferrari’}, \orgname{University of Modena and Reggio Emilia}, \orgaddress{\city{Modena}, \country{Italy}}}

\affil[11]{\orgname{University of Cassino and Southern Lazio}, \orgaddress{\city{Cassino}, \country{Italy}}}


\abstract{

Accurate survival prediction in Non-Small Cell Lung Cancer (NSCLC) requires integrating clinical, radiological, and histopathological data. Multimodal Deep Learning (MDL) can improve precision prognosis, but small cohorts and missing modalities limit its clinical applicability, as conventional approaches enforce complete-case filtering or imputation. We present a missing-aware multimodal survival framework that combines Computed Tomography (CT), Whole-Slide Histopathology Images (WSI), and structured clinical variables for overall survival modeling in unresectable stage II–III NSCLC. 
The framework uses Foundation Models (FMs) for modality-specific feature extraction and a missing-aware encoding strategy that enables intermediate multimodal fusion under naturally incomplete modality profiles. 
By design, the architecture processes all available data without dropping patients during training or inference. 
Intermediate fusion outperforms unimodal baselines and both early and late fusion strategies, with the trimodal configuration reaching a C-index of 74.42. Modality-importance analyses show that the fusion model adapts its reliance on each data stream according to representation informativeness, shaped by the alignment between FM pretraining objectives and the survival task. 
The learned risk scores produce clinically meaningful stratification of disease progression and metastatic risk, with statistically significant log-rank tests across all modality combinations, supporting the translational relevance of the proposed framework.

}

\rr
\bb


\maketitle
\section{Introduction}\label{sec1}

Lung cancer persists as a leading cause of cancer-related mortality worldwide, with Non-Small Cell Lung Cancer (NSCLC) accounting for approximately 85\% of all diagnosed cases~\cite{NSLCadvances}.
The advent of targeted therapy and immunotherapy has markedly reshaped the therapeutic landscape of NSCLC~\cite{Immuno-Targeted}. 
However, the variability in individual treatment effectiveness makes robust risk stratification a pressing clinical need.
Accurate risk estimation is supportive to critical therapeutic choices~\cite{LC_review_imaging}, such as selecting the appropriate treatment modality, adjusting treatment intensity, and defining follow-up strategies, according to each patient’s expected clinical trajectory.
Therapeutic outcomes for non-small cell lung cancer (NSCLC) might not be fully attributable to a single tumor biomarker or image-based diagnostic technique.
This means that understanding a patient's prognostic risk requires information drawn from complementary sources~\cite{guarrasi2025beyond, guarrasi2024multimodal}. 

Histopathology, radiology, genomic information, and Electronic Health Records (EHR) provide multi-scale views of tumor morphology, molecular profile, and patient health status~\cite{Captier2025Integration}.
Histopathology, obtained from digitalized whole slide images (WSI) of the biopsy examination, remains the gold standard for diagnosis, providing high-resolution insight into the microscopic morphology of pathological tissue. 
This examination holds significant prognostic value by revealing the tumor microenvironment and immune infiltration patterns, which are key determinants for survival.
Yet, due to the invasive and focal nature of biopsies, they frequently fail to capture the complete spatial variability of the tumor.~\cite{PATHOLOGY_CANCER}.
Radiological imaging provides a non-invasive, macroscopic view of the entire tumour burden. 
In NSCLC, this role is primarily fulfilled by computed tomography (CT).
Although CT scans overcome the spatially localized and invasive nature of biopsies, they provide a limited survival signal when used in isolation to predict disease outcome~\cite{paverd2024radiology}.
Molecular sequencing provides genomic information that characterizes the tumour's underlying genetic landscape. 
Although the identification of oncogenic drivers and mutations yields critical predictive biomarkers for targeted therapy, genomics alone does not capture the physical properties of the pathology.
Finally, EHRs aggregate the essential patients' clinical context, ranging from physiological parameters and comorbidities to treatment history, into structured tabular data. 
These clinical variables are indispensable for linking biological findings to actual survival outcomes and monitoring disease progression~\cite{COMBINE_MOLECULAR_IMAGING_CLINICAL, RT_comparison}.
To enable robust modeling and harmonization across patients, a common practice is to process EHR data, aggregating key physiological variables, comorbidities, and treatment history, uniforming them into structured tabular formats.
\bb

Faced with this intricate tangle of multi-source information, Artificial Intelligence (AI) research in medicine has converged toward Multimodal Deep Learning (MDL), where multiple modalities are fused to capture a holistic view of the biological mechanisms underlying the disease~\cite{doan2026bridging}.
In NSCLC, prognosis is shaped by factors spanning multiple biological scales, from cellular-level tissue alterations to macroscopic tumour characteristics and broader clinical indicators. 
MDL aims to capture the latent non-linear relationships between these complementary views, i.e., WSIs, CT scans, and clinical data, translating their joint signal into more accurate survival estimation and supporting the transition toward personalized cancer care~\cite{Niu2025Multimodal}.

The integration of these complementary data streams enables clinicians to move beyond a static snapshot of disease status and more accurately anticipate disease progression and patient outcomes.
To translate this multimodal perspective into actionable predictions, prognostic estimation can be formulated as an \textbf{Overall Survival (OS)} prediction task~\cite{machin2006survival}, the standard quantitative framework for estimating patient outcomes~\cite{Soenksen2022Integrated}.
\bb
Unlike binary classifiers, which collapse prognosis into a static mortality label, OS modeling uses time-to-event analysis to estimate survival probabilities over time. 
Handling of censored observations, i.e., patients who are lost to follow-up or who remain event-free at analysis, is a key strength of this formulation, thus ensuring that all available longitudinal information contributes to estimate the survival prediction~\cite{caruso2024deep}.

From these survival functions, OS models generate continuous, patient-specific risk scores that quantify individual treatment trajectories. 
These scores are then used to stratify patients into clinically meaningful risk groups, typically via median or quartile thresholds, to distinguish individuals with aggressive disease trajectories from those who are likely to experience more indolent courses~\cite{GLIO_SURV}. 
Such stratification is central to precision oncology, informing decisions on therapeutic intensity, surveillance frequency, and eligibility for targeted or experimental treatments.

Historically, survival modeling has been largely driven by Machine Learning (ML) approaches, most notably the Cox Proportional Hazards (CPH) model and Random Survival Forests (RSF)~\cite{CPH_RSF_Comparison}. 
While effective on structured, low-dimensional clinical variables, these models are inherently limited in their ability to operate directly on high-dimensional data such as CT volumes or WSIs. 
This limitation has motivated the adoption of MDL as a more suitable framework for accurate and clinically actionable survival estimation in NSCLC~\cite{MMSURV, MM_NSLC, oh2023deep, GLIO_SURV}.
Despite the clear advantages of MDL, its applicability in real clinical scenarios remains fundamentally constrained by the limited availability of comprehensive multimodal datasets. 
In oncological settings, two pervasive structural limitations systematically hinder the robustness and generalizability of MDL models: restricted cohort sizes and the occurrence of partially missing or fragmented modalities~\cite{guarrasi2024multimodal, Yang2024Multimodal}.
First, the inherent scarcity of labeled data combined with the heterogeneity of medical datasets creates a high risk of overfitting, particularly when training imaging encoders, such as Convolutional Neural Networks (CNNs), from scratch on high-dimensional inputs.
To overcome this bottleneck, the field has increasingly adopted Foundation Models (FMs), whose large-scale pre-training enables them to capture generalist representations of the underlying data distribution~\cite{bommasani2021opportunities}. 
This allows the extraction of high-level, semantically rich features from each input modality, reducing the dependency on large task-specific labeled datasets.

In healthcare, FMs have demonstrated remarkable robustness in high-variance clinical settings~\cite{BENCH_FM, li2024artificial}. 
While their utility as feature extractors is well-validated for diagnostic tasks across both CT and WSI images~\cite{conch, titan, MERLIN,  Xu2024A, hamamci2026generalist, pai2025vision}, their application to prognosis prediction remains underexplored~\cite{ruffini2025benchmarking}. 
Unlike static diagnosis, deploying these generalist representations for survival analysis requires validating whether they capture the temporal and prognostic signals hidden within the tumor.

Second, medical data are inherently incomplete, with missingness occurring at the level of both individual variables and entire diagnostic modalities~\cite{NAIM, lipkova2022artificial}. 
Clinical tabular data may be incomplete due to non-response, data entry errors, patient attrition, data corruption, or systematic omissions linked to specific clinical pathways. 
At the modality level, CT scans may be unavailable for patients treated at external institutions, and histopathology slides may be absent when biopsies are not performed or are technically unsuccessful. 
Consequently, real-world cohorts rarely contain a complete and uniform set of measurements across all patients.

On these grounds, this work presents a missing-aware multimodal framework for OS prediction in NSCLC, validated on an internal cohort of unresectable stage II–III patients undergoing radical chemoradiotherapy.
A central aspect of this study is the exploration of Foundation Models as feature extractors for both CT volumes and WSIs within a small clinical cohort, investigating whether generalist representations acquired through large-scale pretraining can effectively substitute for task-specific encoder training in data-scarce oncological settings. 
Building on this premise, the proposed framework comprises three stages: (i) FM-based unimodal feature extraction for CT and WSI modalities, mapping raw imaging data into high-level representations without requiring cohort-specific fine-tuning of vision encoders; (ii) missing-aware representation learning, where each modality is encoded through a transformer-based architecture with an adaptive masking mechanism that enables the network to structurally ignore unavailable inputs; and (iii) an intermediate multimodal fusion strategy, in which the modality-specific latent representations are concatenated and processed by a multimodal head, capturing cross-modal interactions and producing a continuous, patient-specific hazard estimate. 
The framework is resilient to missing modalities by design, allowing both training and inference on incomplete patient profiles without imputation or sample exclusion. 
Additionally, we performed a systematic modality-importance analysis by progressively masking individual data streams at test time and monitoring the corresponding impact on survival prediction, providing an interpretable view of modality relevance and clinically actionable insights on model reliability.

\section{Results}\label{sec2}
This section presents the results of the proposed three-stage, missing-aware multimodal learning framework. 
We report performance across unimodal, bimodal, and trimodal configurations, with a focus on robustness under incomplete patient profiles.

\subsection{Cohort Characteristics}\label{cohort}

% Table 1- Cohort - Part 1
\begin{table*}[!ht]
\tiny
\renewcommand{\arraystretch}{0.85}
\caption{\textbf{Cohort characteristics stratified by survival status.}}
\resizebox{\textwidth}{!}{\begin{tabular}{lccc|c}
\toprule

\textbf{Characteristic} & \textbf{Overall (n=179)} & \textbf{Non-Survivor (n=99)} & \textbf{Survivor (n=80)} & \textbf{P-value} \\
\midrule

\textbf{Age (years)} & 69.8 $\pm$ 12.0 & 70.0 $\pm$ 12.5 & 69.6 $\pm$ 11.3 & 0.820 \\
\midrule
\textbf{Sex} & & & & 0.039$^{*}$ \\
\quad Female & 50 (27.93\%) & 21 (21.2\%) & 29 (36.2\%) & \\
\quad Male   & 129 (72.07\%) & 78 (78.8\%) & 51 (63.7\%) & \\
\midrule
\textbf{Weight (kg)} & 70.3 $\pm$ 13.9 & 72.4 $\pm$ 14.0 & 68.1 $\pm$ 13.5 & 0.160 \\
\midrule
\textbf{Height (cm)} & 168.3 $\pm$ 8.9 & 169.8 $\pm$ 9.1 & 166.7 $\pm$ 8.5 & 0.112 \\
\midrule
\textbf{NRS} & 0.3 $\pm$ 1.3 & 0.4 $\pm$ 1.5 & 0.2 $\pm$ 1.0 & 0.419 \\
\midrule
\textbf{Smoking habitus} & & & & 0.521 \\
\quad Ex-smoker     & 43 (53.75\%) & 23 (56.1\%) & 20 (51.3\%) & \\
\quad Smoker        & 28 (35.00\%) & 15 (36.6\%) & 13 (33.3\%) & \\
\quad Non-smoker    & 9 (11.25\%) & 3 (7.3\%)  & 6 (15.4\%) & \\
\midrule
\textbf{Cigarettes/day} & 22.4 $\pm$ 18.7 & 23.8 $\pm$ 19.5 & 21.0 $\pm$ 18.2 & 0.547 \\
\midrule
\textbf{Stage at diagnosis} & & & & 0.540 \\
\quad IIA   & 2 (1.12\%) & 1 (1.0\%) & 1 (1.2\%) & \\
\quad IIB   & 5 (2.79\%)& 3 (3.0\%) & 2 (2.5\%) & \\
\quad IIIA  & 84 (46.93\%) & 48 (48.5\%) & 36 (45.0\%) & \\
\quad IIIB  & 71 (39.66\%) & 41 (41.4\%) & 30 (37.5\%) & \\
\quad IIIC  & 17 (9.50\%)& 6 (6.1\%)  & 11 (13.8\%) & \\
\midrule

\textbf{Histology} & & & & 0.331 \\
\quad Adenocarcinoma & 90 (50.28\%) & 48 (48.5\%) & 42 (52.5\%) & \\
\quad Squamous carcinoma & 80 (44.69\%) & 44 (44.4\%) & 36 (45.0\%) & \\
\quad NOS & 5 (2.79\%) & 3 (3.0\%) & 2 (2.5\%) & \\
\quad Other & 4 (2.23\%) & 4 (4.0\%) & 0 (0.0\%) & \\
\midrule

\textbf{Comorbidity 1} & & & & 0.479 \\
\quad Vascular & 48 (64.00\%) & 25 (65.8\%) & 23 (62.2\%) & \\
\quad Metabolic & 12 (16.00\%) & 4 (10.5\%) & 8 (21.6\%) & \\
\quad Renal & 2 (2.67\%) & 2 (5.3\%) & 0 (0.0\%) & \\
\quad Other & 11 (14.67\%) & 6 (15.8\%) & 5 (13.5\%) & \\
\quad Pulmonary & 2 (2.67\%) & 1 (2.6\%) & 1 (2.7\%) & \\
\midrule

\textbf{Comorbidity 2} & & & & 0.756 \\
\quad Vascular & 6 (11.76\%) & 2 (7.7\%) & 4 (16.0\%) & \\
\quad Metabolic & 26 (50.98\%) & 15 (57.7\%) & 11 (44.0\%) & \\
\quad Renal & 3 (5.88\%) & 2 (7.7\%) & 1 (4.0\%) & \\
\quad Pulmonary & 5 (9.80\%) & 2 (7.7\%) & 3 (12.0\%) & \\
\quad Other & 11 (21.57\%) & 5 (19.2\%) & 6 (24.0\%) & \\
\midrule

\textbf{Comorbidity 3} & & & & 0.253 \\
\quad Vascular & 1 (6.67\%) & 0 (0.0\%) & 1 (12.5\%) & \\
\quad Metabolic & 1 (6.67\%) & 1 (14.3\%) & 0 (0.0\%) & \\
\quad Renal & 2 (13.33\%) & 0 (0.0\%) & 2 (25.0\%) & \\
\quad Other & 10 (66.67\%) & 6 (85.7\%) & 4 (50.0\%) & \\
\quad Pulmonary & 1 (6.67\%) & 0 (0.0\%) & 1 (12.5\%) & \\
\midrule
\textbf{ECOG PS} & & & & 0.101 \\
\quad 0 & 60 (70.59\%) & 26 (60.5\%) & 34 (81.0\%) & \\
\quad 1 & 23 (27.06\%) & 16 (37.2\%) & 7 (16.7\%) & \\
\quad 2 & 2 (2.35\%) & 1 (2.3\%) & 1 (2.4\%) & \\
\bottomrule
\end{tabular}}
\parbox{\textwidth}{\raggedright
\scriptsize{
Continuous variables are presented as mean $\pm$ standard deviation, and categorical variables as n ($\%$). 
P-values were calculated using one-way analysis of variance for continuous variables and chi-square test for categorical variables. 
RT \textit{RadioTherapy}, Ch.T \textit{ChemoTherapy}. 
Statistical significance is indicated as follows: $p < 0.05$ (*), $p < 0.005$ (**), and $p < 0.001$ (***).
}
}
\label{tab:baseline_characteristics_part1}
\end{table*}

% Table 1 - Part 2
\begin{table*}[!ht]
\centering
\tiny
\renewcommand{\arraystretch}{0.85}
\caption{\textbf{(Table 1 Continued) Cohort characteristics stratified by survival status.}}
\resizebox{\textwidth}{!}{
\begin{tabular}{lccc|c}
\toprule
\textbf{Characteristic} & \textbf{Overall (n=179)} & \textbf{Non-Survivor (n=99)} & \textbf{Survivor (n=80)} & \textbf{P-value} \\
\midrule
\textbf{EGFR mutation} & & & & $0.033^{*}$ \\
\quad Negative & 41 (87.23\%) & 23 (100.0\%) & 18 (75.0\%) & \\
\quad Positive & 6 (12.77\%) & 0 (0.0\%) & 6 (25.0\%) & \\
\midrule

\textbf{ALK rearrangement} & & & & 0.233 \\
\quad Negative & 45 (93.75\%) & 24 (100.0\%) & 21 (87.5\%) & \\
\quad Positive & 3 (6.25\%) & 0 (0.0\%) & 3 (12.5\%) & \\
\midrule

\textbf{MET alteration} & & & & 1.000 \\
\quad Negative & 4 (66.67\%) & 1 (100.0\%) & 3 (60.0\%) & \\
\quad Positive & 2 (33.33\%) & 0 (0.0\%) & 2 (40.0\%) & \\
\midrule
\textbf{PD-L1 (\%)} & 0.2 $\pm$ 0.3 & 0.2 $\pm$ 0.3 & 0.2 $\pm$ 0.3 & 0.776 \\
\midrule

\textbf{RT dose (Gy)} & 59.1 $\pm$ 8.9 & 58.3 $\pm$ 10.3 & 60.1 $\pm$ 6.8 & 0.176 \\
\midrule
\textbf{Number of fractions} & 28.6 $\pm$ 5.3 & 27.9 $\pm$ 6.3 & 29.3 $\pm$ 3.8 & 0.211 \\
\midrule
\textbf{RT Technique} & & & & $<0.001^{***}$ \\
\quad  3DCRT & 43 (30.28\%) & 32 (43.2\%) & 11 (16.2\%) & \\
\quad  VMAT & 80 (56.34\%) & 39 (52.7\%) & 41 (60.3\%) & \\
\quad  Mixed & 6 (4.23\%) & 2 (2.7\%) & 4 (5.9\%) & \\
\quad  RA & 13 (9.15\%) & 1 (1.4\%) & 12 (17.6\%) & \\
\midrule

\textbf{Induction Ch.T} & & & & 0.465 \\
\quad CC & 13 (43.33\%) & 7 (50.0\%) & 6 (37.5\%) & \\
\quad gem & 1 (3.33\%) & 0 (0.0\%) & 1 (6.2\%) & \\
\quad C & 15 (50.00\%) & 6 (42.9\%) & 9 (56.2\%) & \\
\quad CC-gem & 1 (3.33\%) & 1 (7.1\%) & 0 (0.0\%) & \\
\midrule

\textbf{Concomitant Ch.T} & & & & 0.351 \\
\quad CC & 58 (44.96\%) & 31 (42.5\%) & 27 (48.2\%) & \\
\quad Gem. & 13 (10.08\%) & 8 (11.0\%) & 5 (8.9\%) & \\
\quad C & 8 (6.20\%) & 3 (4.1\%) & 5 (8.9\%) & \\
\quad CC+Gem & 35 (27.13\%) & 22 (30.1\%) & 13 (23.2\%) & \\
\quad Al. & 2 (1.55\%) & 0 (0.0\%) & 2 (3.6\%) & \\
\quad other & 13 (10.08\%) & 9 (12.3\%) & 4 (7.1\%) & \\
\midrule

\textbf{RT (days)} & 48.7 $\pm$ 12.2 & 48.3 $\pm$ 11.5 & 49.1 $\pm$ 13.0 & 0.695 \\
\midrule
\textbf{Suspension days} & 2.9 $\pm$ 7.6 & 3.8 $\pm$ 9.9 & 2.0 $\pm$ 4.2 & 0.279 \\
\midrule
\textbf{Adjuvant immunotherapy} & & & & 0.010$^{*}$ \\

\quad No & 138 (77.09\%) & 84 (84.8\%) & 54 (67.5\%) & \\
\quad Yes & 41 (22.91\%) & 15 (15.2\%) & 26 (32.5\%) & \\
\midrule
\textbf{Esophageal toxicity} & & & & 0.183 \\
\quad Grade 0 & 57 (67.86\%) & 31 (72.1\%) & 26 (63.4\%) & \\
\quad Grade $>$0 & 27 (32.14\%) & 12 (27.9\%) & 15 (36.6\%) & \\
\midrule
\textbf{Pulmonary toxicity} & & & & 0.250 \\
\quad Grade 0 & 75 (88.24\%) & 36 (83.7\%) & 39 (92.9\%) & \\
\quad Grade $>$0 & 10 (11.76\%) & 7 (16.3\%) & 3 (7.1\%) & \\
\midrule
\textbf{Hemoglobin toxicity} & & & & 0.276 \\
\quad Grade 0 & 76 (89.41\%) & 36 (83.7\%) & 40 (95.2\%) & \\
\quad Grade $>$0 & 9 (10.59\%) & 7 (16.3\%) & 2 (4.8\%) & \\
\midrule
\textbf{Neutrophil toxicity} & & & & 0.709 \\
\quad Grade 0 & 72 (84.71\%) & 37 (86.0\%) & 35 (83.3\%) & \\
\quad Grade $>$0 & 13 (15.29\%) & 6 (14.0\%) & 7 (16.7\%) & \\
\midrule
\textbf{Platelet toxicity} & & & & 0.440 \\
\quad Grade 0 & 68 (80.00\%) & 36 (83.7\%) & 32 (76.2\%) & \\
\quad Grade $>$0 & 17 (20.00\%) & 7 (16.3\%) & 10 (23.8\%) & \\
\midrule
\textbf{PFS event} & & & & $<0.001^{***}$ \\
\quad censored & 65 (36.31\%)   &  21 (21.2\%) & 44 (55.0\%)  & \\
\quad uncensored & 114 (63.69\%) & 78 (78.8\%) & 36 (45.0\%)  & \\
\midrule
\textbf{DM event} & & & & $<0.001^{***}$ \\
\quad censored   &  88 (49.16\%)  & 35 (35.4\%) & 53 (66.2\%)  & \\
\quad uncensored &  91 (50.84\%)  & 64 (64.6\%) & 27 (33.8\%)  & \\

\bottomrule
\end{tabular}}
\parbox{\textwidth}{\raggedright
\scriptsize{
Continuous variables are presented as mean $\pm$ standard deviation, and categorical variables as n ($\%$). 
P-values were calculated using one-way analysis of variance for continuous variables and chi-square test for categorical variables. 
RT \textit{RadioTherapy}; Ch.T \textit{ChemoTherapy}; PFS \textit{Progression Free Survival}; DM \textit{Distant Metastasis}. 
Statistical significance is indicated as follows: $p < 0.05$ (*), $p < 0.005$ (**), and $p < 0.001$ (***).
}
}

\label{tab:baseline_characteristics_part2}
\end{table*}

Our study is based on a retrospective cohort of 179 patients with unresectable Stage IIA, IIB, IIIA, IIIB and IIIC NSCLC treated with radical Chemo-Radiotherapy.
Tables~\ref{tab:baseline_characteristics_part1} and~\ref{tab:baseline_characteristics_part2} present a univariate analysis of baseline cohort characteristics in relation to 5-year overall survival status. Statistical significance was evaluated using one-way analysis of variance (ANOVA) for continuous variables and the $\chi^2$ test for categorical variables.

Patients are stratified into Survivors ($n=80$) and Non-survivors ($n=99$) based on their 5-year survival status.
While the primary endpoint is time-to-event overall survival, this univariate stratification serves as an exploratory characterization of the cohort, offering an interpretable summary of baseline differences and confirming that the observed covariate trends align with the survival outcome.

Across demographic variables, the mean age of the cohort was $69.8 \pm 12.0$ years, with similar distributions in the Survivor ($69.6 \pm 11.3$) and Non-survivor ($70.0 \pm 12.5$) groups ($p=0.820$). 
A significant difference was observed for sex ($p=0.039$), with males representing $78/99$ ($78.8\%$) of non-survivors compared to $51/80$ ($63.7\%$) of survivors, indicating a statistically significant association between sex and worse survival outcomes.
Stage at diagnosis was similarly distributed ($p=0.540$), with stage~IIIA observed in $48/99$ non-survivors and $36/80$ survivors, and Stage~IIIB in $41/99$ and $30/80$ patients, respectively.
Histological subtype stratification were comparable between the groups ($p=0.331$), with majority of cases identified as adenocarcinoma ($48/99$ for non survivor compared to $42/80$ for survivor) and squamous carcinoma ($44/99$ and $36/80$). 
Similar patterns were observed for renal, pulmonary, and other comorbidity categories.
For example, vascular comorbidities were reported in $25/38$ ($65.8\%$) non-survivors and $23/37$ ($62.2\%$) survivors ($p=0.479$), and metabolic comorbidities in $15/26$ ($57.7\%$) and $11/25$ ($44.0\%$) patients ($p=0.756$). 
Among molecular biomarkers, EGFR mutation status differed between groups ($p=0.033$), with positive cases found exclusively among Survivors ($7.5\%$ compared to $0\%$ in Non-survivors). 
This difference highlights the known favorable impact of EGFR mutations, likely attributable to the distinct biology of this subtype and the availability of effective targeted therapies~\cite{Lee2013Impact}.
Moreover, a highly significant variation was also observed in radiotherapy techniques ($p<0.001$), reflecting the survival advantage associated with modern delivery methods. 
Survivors were more frequently treated with VMAT ($60.3\%$), whereas usage of 3D-CRT was nearly three times higher in the non-survivor group ($43.2\%$ vs. $16.2\%$), suggesting an association between older conformal techniques and poorer prognostic outcomes.

The administration of adjuvant immunotherapy was significantly associated with improved survival ($p=0.010$). 
Its higher prevalence among Survivors ($32.5\%$ vs. $15.2\%$) reflects the benefit of consolidation checkpoint inhibition in Stage III NSCLC, reinforcing that access to contemporary systemic therapy remains a major prognostic determinant in this setting~\cite{pacific_trial}.

However, a dichotomy emerged when analyzing the intermediate clinical endpoints, highlighting that mortality is fundamentally driven by the disease's trajectory rather than baseline demographics.
\textit{Progression-Free Survival (PFS)} events were strongly correlated with mortality ($p < 0.001^{***}$): the vast majority of non-survivors ($78.8\%$) experienced a progression event compared to less than half of the survivors ($45.0\%$).
This pattern was further reinforced by the significantly higher rate of \textit{Distant Metastasis (DM)} in the non-survivor group ($64.6\%$ vs. $33.8\%$, $p < 0.001^{***}$), indicating systemic failure as a hallmark of poor prognosis.
These findings confirm that while the two groups were demographically and clinically balanced at baseline, their outcomes were determined by the distinct biological aggressiveness of the tumor, manifesting as therapeutic resistance (PFS events) and systemic spread (DM).
We examine in detail the impact of progression events on model performance in the following results sections.
Finally, no statistically relevant differences were observed for the remaining baseline characteristics. 
Variables such as age ($p=0.820$), TNM stage ($p=0.540$), histology ($p=0.331$), and ECOG performance status ($p=0.101$) were evenly distributed between survivors and non-survivors groups.

\bb

\subsection{Missing Modalities and Tabular Data Sparsity}\label{sec-missing}

%%%%%% ---------------------------------------------------------------------- ----------------------------------------------------------------------
%%%%%% ---------------------------------------------------------------------- ----------------------------------------------------------------------
%%%%%% MISSING MODALITIES PLOT

\begin{figure*}[!htbp]
    \centering
     \includegraphics[width=\textwidth]{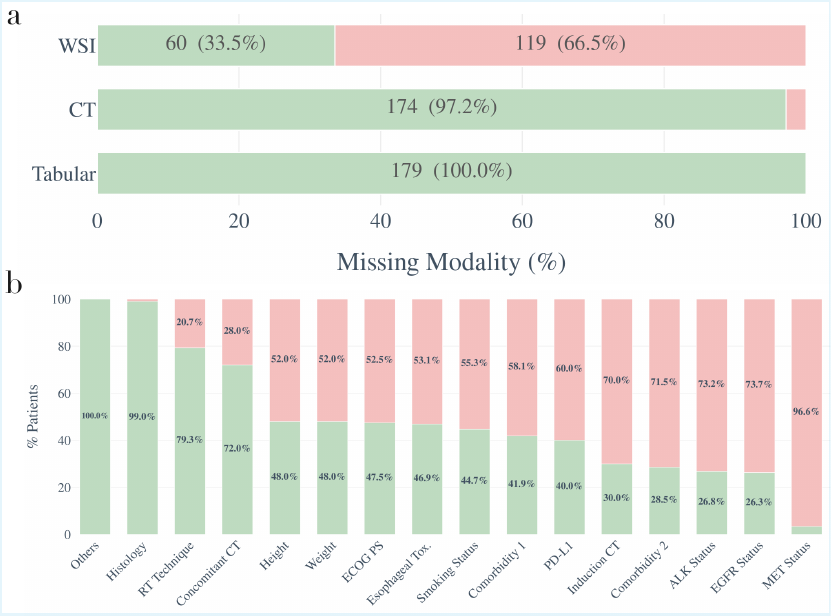}
    \caption{Green indicates the proportion of available data, whereas red denotes missingness.
Panel (a) reports the availability of each modality (WSI, CT, and tabular data) across patients.
Panel (b) illustrates feature-wise missingness within the tabular modality, highlighting the proportion of absent values for each clinical variable.
}
    \label{fig:missingness_combined}
\end{figure*}

%%%%%% 
The cohort in this study is characterized by significant data heterogeneity, presenting variable levels of missingness and feature sparsity.
As displayed in Figure~\ref{fig:missingness_combined}, data availability is unevenly distributed across acquisition modalities and within the tabular clinical domain itself.
At the modality level (Figure~\ref{fig:missingness_combined}.a), WSI constitutes the primary bottleneck, being available only for $33.5\%$ of the cohort ($n=60$). 
In contrast, radiological imaging (CT) and the tabular modality show near-complete coverage, with only $2.8\%$ of cases lacking a scan and $0\%$ lacking the collection of at least the basic informations. 
Nevertheless, the intersection of modalities remains narrow: 22.9\% of patients ($n=41$) possess the full trimodal profile (CT + WSI + Tabular). 
By contrast, radiological imaging (CT) and clinical data (Tabular) demonstrate near-complete availability, with only $2.8\%$ of cases missing a CT scan and no cases lacking baseline patient information.

However, full modality-level availability in the Tabular modality does not imply that its constituent clinical features have been fully collected.
Figure~\ref{fig:missingness_combined}.b shows that the structured clinical data, despite being technically present for all patients, exhibits substantial internal sparsity. 
Several clinically relevant molecular biomarkers, such as EGFR, ALK, and MET mutation status, are absent in $73.7\%$, $73.2\%$, and $96.6\%$ of patients, respectively. 
The absence of these molecular features reflects a structural limitation of standard clinical workflows, where molecular testing is not systematically performed and retrospective retrieval of these results is often challenging or unfeasible.
Laboratory and histopathological descriptors (e.g. PD-L1 expression) demonstrate analogous levels of incompleteness for analogous reasons.
Beyond genomic and histopathological descriptors, clinical information as comorbidities are often incompletely documented, leading to substantial under-reporting of cardiovascular, metabolic, and respiratory conditions. 
Cross-modal and intra-modal sparsity have direct consequences for learning. Conventional multimodal architectures enforce a complete-case design: all modalities must be present, or the patient is imputed or excluded. 

In our cohort, this constraint would remove 77.1\% of cases (Figure~\ref{fig:missingness_combined}a), comprising patients with incomplete diagnostic workups, comorbidities, or limited access to comprehensive staging procedures.
Critically, missing modalities may reflect disease severity itself. 
Patients with the most aggressive trajectories are plausibly those for whom full staging was never completed, meaning their exclusion would systematically deplete the cohort of its hardest and most informative cases.
Selecting only complete cases introduces systematic bias and erodes both sample size and model generalizability.
Our architecture admits the full cohort by design, handling missing modalities and clinical features without imputation or exclusion.

\begin{figure}[!h]
    \centering
    \includegraphics[width=\textwidth]{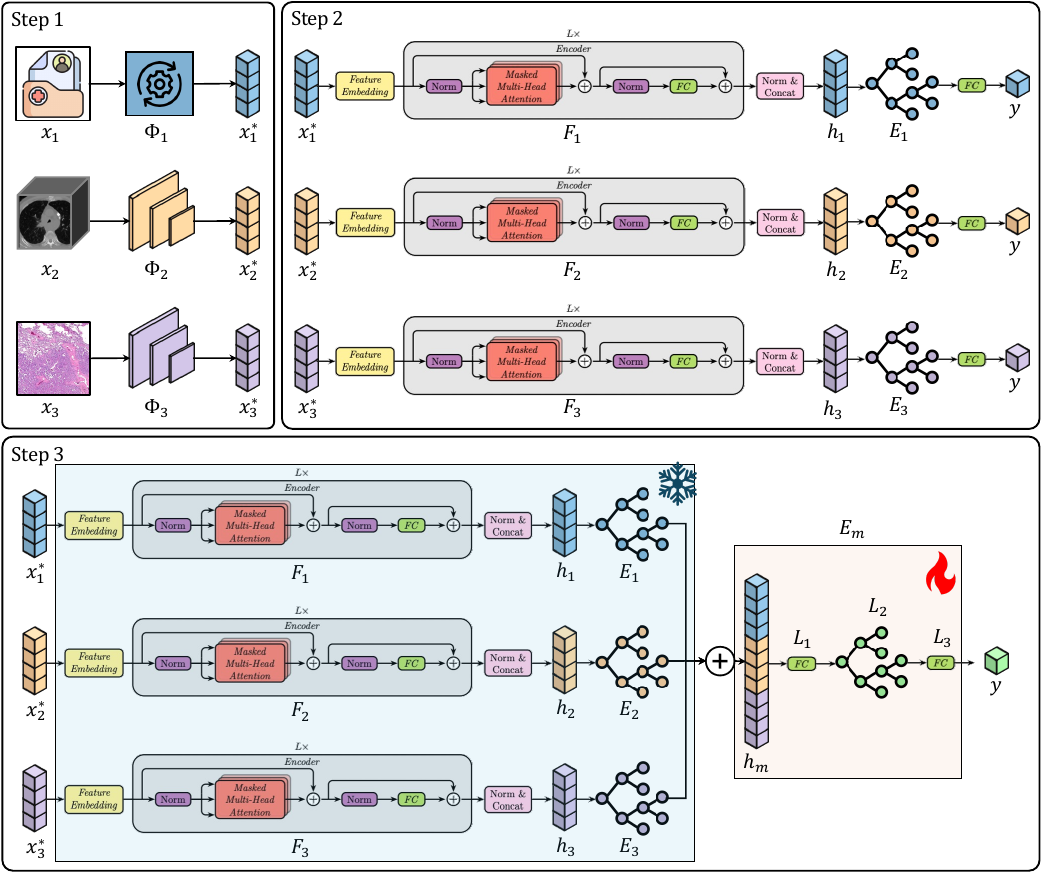}
    \caption{The framework integrates three clinical data streams: Pathology (WSI), Radiology (CT), and Tabular data. 
\textbf{Step-1} extracts fixed embeddings via domain-specific foundation models. 
\textbf{Step-2} processes each modality through a dedicated \texttt{NAIM+ODST} encoder with built-in missing modality handling. 
\textbf{Step-3} implements \texttt{ConcatODST}, where unimodal latent representations are concatenated and passed through a multimodal module to predict the patient-specific hazard $y$.}
    \label{fig:method_architecture}
\end{figure}

\subsection{Missing-Aware Multimodal Survival Modeling}

To support multimodal survival prediction under realistic clinical missingness, we propose a three-stage framework that decouples FM feature extraction, missing-aware unimodal encoding, and intermediate multimodal fusion. 
Figure~\ref{fig:method_architecture} illustrates the complete pipeline and a comprehensive methodological presentation is provided in Section~\ref{sec:methods}.

\noindent 
\bl
\textbf{Step-1} performs modality-specific pre-processing and feature extraction through functions $\Phi_i(\cdot)$, mapping each raw input $x_i$ to a fixed embedding $x_i^\ast$. $\Phi_1$ denotes the tabular pre-processing pipeline, $\Phi_2$ encodes CT volumes via CT-FM~\cite{pai2025vision}, and $\Phi_3$ 
implements a multi-stage WSI pipeline of dedicated foundation models for tissue segmentation, patch encoding, and multi-slide aggregation. 
\bb

\noindent 
\textbf{Step-2} transforms each $x_i^\ast$ through a modality-specific encoder $F_i(\cdot)$ with missing-aware tokenization and masked self-attention, yielding a latent representation $h_i$ conditioned only on observed inputs. Each $h_i$  is then processed by an \textsc{ODST} layer $E_i(\cdot)$ and a fully connected projection to produce a unimodal risk estimate.
\bl

\noindent 
\textbf{Step-3} implements \textsc{ConcatODST}, the proposed intermediate fusion model. 
The unimodal encoders are loaded from Step 2 and kept frozen. 
The latent vectors $h_i$ are concatenated ($\oplus$) into a unified representation $h_m$, passed through the multimodal module $E_m(\cdot)$, composed of a fully connected layer downscaling to $1/3$ of the input dimension followed by an \textsc{ODST} layer, and a final fully connected layer that generates the multimodal survival risk prediction $y$. 
The ablation study on the choice of $E_m(\cdot)$ is detailed in the supplementary material (Section S.1). 
The comparison study for the selection of the CT-FM foundation model for the CT modality is detailed in the supplementary material (Section S.3)
\bb

\subsection{Prognostic Separability in FMs' Representation Spaces}\label{sec-umap-fm}

%%%%%% ---------------------------------------------------------------------- ----------------------------------------------------------------------
% EMBEDDING PLOT
\begin{figure*}[!htbp]
 \centering
 \includegraphics[width=\textwidth]{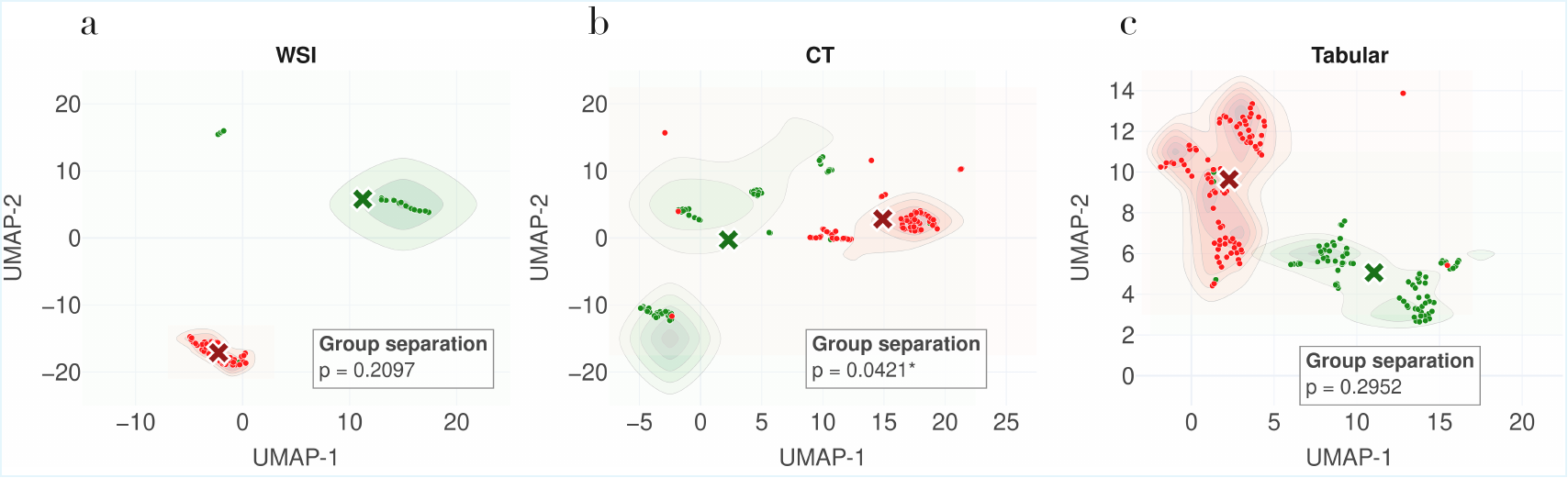}
\caption{Two-dimensional UMAP projections of foundation model embeddings for WSI (a), CT (b), and tabular data (c).
Each plot shows the distribution of survivors (green) and non-survivors (red), with kernel density contours and class centroids.
Reported p-values quantify group separation in the embedding space.}
\label{fig:umap_modalities}
\end{figure*}
%%%%%% ---------------------------------------------------------------------- ----------------------------------------------------------------------
%%%%%% ---------------------------------------------------------------------- ----------------------------------------------------------------------
%%%%%% ---------------------------------------------------------------------- ----------------------------------------------------------------------
\bb
To enable unimodal representation learning under limited cohort size, we extracted imaging features from pretrained foundation models (FMs).
For WSI, the multi-instance nature of WSI images requires a two-stage pipeline: patch-level feature extraction followed by aggregation into a single slide-level embedding. 
Therefore, We combined CLAM-based patch aggregation with MI-Zero vision encoders and the TITAN framework~\cite{titan, conch, CLAM}, producing a $768$-dimensional representation.
For CT volumes, vision transformer based FMs pretrained on large-scale volumetric datasets expose a class token that serves directly as a compressed scan-level embedding, reducing extraction to a single forward pass. 
Selecting among available CT-specific FMs was less immediate, as available FMs for CT differ substantially in pretraining strategy and output dimensionality.
We therefore conducted a comparative analysis across CT-CLIP~\cite{hamamci2026generalist}, MERLIN~\cite{MERLIN}, and CT-FM~\cite{pai2025vision}; the comparison is detailed in the supplementary material (Section S.3). 
Based on this comparison, we selected CT-FM~\cite{pai2025vision}, which outputs a 512-dimensional token per volume.
\bb
For completeness, tabular variables were included after imputing missing values using median imputation for continuous features and $k$-nearest neighbors for categorical features.

Using these embeddings, we ran a UMAP-based representation analysis to probe the prognostic signal encoded by each modality (Figure~\ref{fig:umap_modalities}). 
This representation analysis serves as a preliminary, mortality-based assessment rather than a direct reflection of the time-to-event task.
Patients are stratified into \textit{Survivors} and \textit{Non-Survivors} by 5-year status, mirroring the cohort characterization in Section~\ref{cohort}. 
WSI embeddings (Fig.~\ref{fig:umap_modalities}.a) show locally coherent substructures, likely reflecting histological tissue organization, yet without statistically significant outcome-level separation ($p = 0.2097$).
The unsupervised histopathological representations encode meaningful morphological patterns but do not linearly discriminate prognostic groups, plausibly due to high intra-class heterogeneity and the limited number of available WSIs for this study.
Differently, CT embeddings (Fig.~\ref{fig:umap_modalities}.b) exhibit a statistically robust linear separation between \textit{Survivors} (green) and \textit{Non-Survivors} (red), confirmed by a significant group-separation test ($p = 0.0421$).
The tabular feature space (Fig.~\ref{fig:umap_modalities}.c) shows high variance and no evident outcome-group separation ($p = 0.2952$), confirming that clinical variables alone are insufficient to stratify patients in this cohort.
\bb
Taken together, these observations motivate a multimodal integration strategy, as none of the three data streams, when considered in isolation, provides a fully reliable prognostic separation, yet they capture complementary signals that may become synergistic when jointly modeled. 
In the following section, we move beyond exploratory mortality-based analyses and quantitatively evaluate the predictive performance of each unimodal representation on the full survival modeling task.
\bb

\subsection{
FMs enable Unimodal Survival Modeling}\label{sec-fm}
%%%%%% ---------------------------------------------------------------------- 
%%%%%% ---------------------------------------------------------------------- 
%%%%%% ---------------------------------------------------------------------- 
% UNIMODAL TABLE

% Unimodal table performances
\begin{table*}[!htbp]
\centering
\small
\renewcommand{\arraystretch}{1}
\setlength{\tabcolsep}{2pt}
\label{tab:unimodal_merged}
\setlength{\tabcolsep}{5pt}
\resizebox{\textwidth}{!}{
\begin{tabular}{llcccc}
\toprule
\textbf{Modality} & \textbf{Model} & \textbf{Type} & \textbf{C-index}  & \textbf{td-AUC} & \textbf{Uno}\\
\midrule
% ===================== WSI ======================
\multirow{7}{*}{\texttt{WSI}}
 & \texttt{CPH}       & \texttt{ML} & 49.71 \tiny{± 2.27} & 43.79 \tiny{± 3.29} & 50.62 \tiny{± 2.26} \\
 & \texttt{RSF}       & \texttt{ML} & \textbf{65.47 \tiny{± 1.15}}  &  \textbf{62.68 \tiny{± 2.69}}& \textbf{62.51 \tiny{± 1.67}} \\
 & \texttt{SGB}       & \texttt{ML} & 55.57 \tiny{± 4.16}  & 53.95 \tiny{± 7.61} & 52.55 \tiny{± 5.62}\\
\cmidrule{2-6}
 & \texttt{NN       } & \texttt{DL} & 47.85 \tiny{$\pm$ 6.47} & 41.64 \tiny{$\pm$ 5.95} & 45.21 \tiny{$\pm$ 5.49} \\
 & \texttt{NAIM     } & \texttt{DL} & 55.14 \tiny{$\pm$ 6.86} & 48.67 \tiny{$\pm$ 7.18} & 52.55 \tiny{$\pm$ 6.09} \\
 & \texttt{NAIM+ODST} & \texttt{DL} & \underline{$58.81$ \tiny{$\pm$ 4.04}} & \underline{$55.33$ \tiny{$\pm$ 4.65}} & \underline{$56.28$ \tiny{$\pm$ 2.69}} \\
\midrule[1.5pt]

\multirow{7}{*}{\texttt{CT}}
& \texttt{CPH}       & \texttt{ML} & 58.13 \tiny{± 2.51} & 60.78 \tiny{± 2.28} & 56.88 \tiny{± 2.12} \\
& \texttt{RSF}       & \texttt{ML} & 66.62 \tiny{± 6.09} & 71.90 \tiny{± 7.71} & 61.38 \tiny{± 5.15} \\
& \texttt{SGB}       & \texttt{ML} & \textbf{70.22 \tiny{± 5.05}} &  \underline{74.48 \tiny{± 5.76}} & \textbf{66.24 \tiny{± 4.37}} \\
\cmidrule{2-6}
 & \texttt{NN       } & \texttt{DL} & 59.98 \tiny{± 4.24} & 59.39 \tiny{± 3.84} & 58.63 \tiny{± 4.81} \\
 & \texttt{NAIM     } & \texttt{DL} & 63.39 \tiny{± 2.25} & 61.57 \tiny{± 2.16} & 61.23 \tiny{± 3.48} \\
 & \texttt{NAIM+ODST} & \texttt{DL} &  \underline{68.75 \tiny{± 5.29}} & \textbf{74.57 \tiny{± 6.42}} &  \underline{64.42 \tiny{± 4.86}} \\
\midrule[1.5pt]

% ===================== Tabular ======================
\multirow{7}{*}{\texttt{Tabular}}
 & \texttt{CPH}       & \texttt{ML} & 59.72 \tiny{± 4.78} & 65.05 \tiny{± 7.39} & 57.75 \tiny{± 3.39} \\
 & \texttt{RSF}       & \texttt{ML} & \textbf{67.47 \tiny{± 3.61}} & \textbf{75.28 \tiny{± 3.84}} & \underline{62.39 \tiny{± 2.20}}  \\
 & \texttt{SGB}       & \texttt{ML} & 57.58 \tiny{± 4.12} & 60.75 \tiny{± 5.36} & 56.72 \tiny{± 3.95}  \\
\cmidrule{2-6}
 & \texttt{NN}        & \texttt{DL} & $60.67$ \tiny{$\pm$ 4.12} & $66.48$ \tiny{$\pm$ 5.82} & $55.94$ \tiny{$\pm$ 2.27} \\
 & \texttt{NAIM}      & \texttt{DL} & $63.85$ \tiny{$\pm$ 3.76} & $69.68$ \tiny{$\pm$ 4.60} & $61.75$ \tiny{$\pm$ 3.33} \\
 & \texttt{NAIM+ODST} & \texttt{DL} & \underline{$65.52$ \tiny{$\pm$ 4.45}} & \underline{$72.73$ \tiny{$\pm$ 5.57}} & \textbf{$\mathbf{63.56}$ \tiny{$\pm$ 4.35}} \\
\botrule
\end{tabular}
}
\caption{Unimodal survival performance across WSI, CT, and Tabular modalities for all evaluated models: three ML baselines (\textsc{CPH}, \textsc{RSF}, \textsc{SGB}) and three DL architectures (\textsc{NN}, \textsc{NAIM}, \textsc{NAIM+ODST}). 
Results are reported as mean $\pm$ \textsc{SE} over 5-fold cross-validation. 
\textbf{Bold} and \underline{underline} indicate the best and second-best score per modality, respectively.}
\end{table*}

%%%%%% ---------------------------------------------------------------------- 
%%%%%% ---------------------------------------------------------------------- 
%%%%%% ---------------------------------------------------------------------- 

Each unimodal representation was evaluated on the time-to-event OS 
task under three ML baselines: \texttt{CPH}~\cite{K2024Survival}, 
\texttt{RSF}~\cite{Ishwaran2008RSF}, and \textsc{\texttt{SGB}}~\cite{Hothorn2006SGB}. 
Regarding the Tabular modality, the three survival models differ in their treatment of missing values. 
These methods differ in how they handle missing tabular features. 
\texttt{CPH} and \texttt{RSF} by design handle missing entries during model fitting and are therefore applied directly to the available clinical features. 
In contrast, \textsc{\texttt{SGB}} requires complete inputs and was trained on KNN-imputed data. 
For imaging modalities, patients without a CT scan or WSI were excluded from that unimodal analysis only. 
Cross-validation splits were kept fixed across all experiments; reported performance thus varies in effective sample size across modalities, reflecting each modality's coverage in the cohort (Section~\ref{sec-missing}).

Three DL architectures were considered alongside the ML baselines. 
We first considered a transformer-based model inspired by the Not-Another-Imputation-Method (\texttt{NAIM})~\cite{NAIM} which addresses missing values through masked self-attention, operating directly on incomplete inputs without imputation. 
The NN baseline shares \texttt{NAIM}'s continuous and categorical embedding scheme but removes the transformer layers, replacing it with a standard MLP with KNN-imputed inputs. 
The NN model serves as a vanilla baseline to isolate the contribution of \texttt{NAIM}'s missing-aware attention mechanism.
Lastly, \texttt{NAIM+ODST} retains the full \texttt{NAIM} encoder but replaces the linear survival head with an Oblivious Differentiable Sparsemax Trees (ODST)~\cite{popov2019neural}, targeting higher-order non-linear interactions between the learned representations.
More implementations details for the ODST hyperparameters are reported in the supplementary materials (Section S.2).

\bl
Table~\ref{tab:unimodal_merged} reports unimodal results across all six models under 5-fold cross-validation, using C-index, Uno C-index, and td-AUC.
For WSI, the high dimensionality of the FM embeddings challenged all the selected models. 
Among them, \texttt{RSF} achieved the strongest overall ranking (C-index $65.47$, Uno $62.51$), while the NN failed to generalize, yielding poor results across the board (Uno $45.21$, td-AUC $41.64$). 
Lastly, \texttt{NAIM+ODST} reaches the highest td-AUC among DL models ($55.33$) and Uno c-index ($56.28$).
For CT features extracted via CT-FM, \texttt{SGB} led ML models (C-index $70.22$, Uno $66.24$). 
Among DL models, \texttt{NAIM+ODST} achieved the highest td-AUC ($74.57$) across all the models, yielding the second highest C-index ($68.75$), outperforming both \texttt{NAIM} ($63.39$) and the NN ($59.98$).
For Tabular features, \texttt{RSF} was the strongest model (C-index $67.47$, td-AUC $75.28$). 
Similar to the result on the CT, \texttt{NAIM+ODST} obtained the second best score behind \texttt{RSF} on C-index ($65.52$) and surpassed it on Uno ($63.56$ vs $62.39$), while improving over the standalone \texttt{NAIM} encoder by $3.0$ points in td-AUC ($72.73$ vs $69.68$).
\bb
Based on this consistent performance across disparate modalities, \texttt{NAIM+ODST} was selected as the primary unimodal DL encoder for this study. 
This model serves as the unified backbone for mapping both structured clinical features and high-dimensional imaging representations (CT and WSI) into a shared prognostic space.

\subsection{Intermediate Fusion Strategies Maximize Multimodal Synergy}\label{sec-intermediate}

%%%%%% ---------------------------------------------------------------------- ----------------------------------------------------------------------
%%%%%% ---------------------------------------------------------------------- ----------------------------------------------------------------------
%%%%%% ---------------------------------------------------------------------- ----------------------------------------------------------------------

% Unimodal table performances
\begin{table*}[ht]
\centering
\small
\renewcommand{\arraystretch}{0.85}
\setlength{\tabcolsep}{5pt}
\resizebox{\textwidth}{!}{
\begin{tabular}{c|ll|ccc}
\toprule
\textbf{Fusion} & \textbf{Modalities} & \textbf{Model} & \textbf{C-ix} & \textbf{td-AUC} & \textbf{Uno} \\
\midrule
% ======================= Unimodal =======================
\multirow{6}{*}{\rotatebox[origin=c]{0}{\scriptsize \texttt{Unimodal}}}
& \texttt{WSI} &  \texttt{RSF}     & 65.47 \tiny{± 1.15}  &  62.68 \tiny{± 2.69}& 62.51 \tiny{± 1.67} \\
& \texttt{WSI} & \texttt{NAIM+ODST} & $58.81$ \tiny{$\pm$ 4.04} & $55.33$ \tiny{$\pm$ 4.65} & $56.28$ \tiny{$\pm$ 2.69} \\
& \texttt{CT} & \texttt{SGB}   & 70.22 \tiny{± 5.05} &  74.48 \tiny{± 5.76} & 66.24 \tiny{± 4.37} \\
& \texttt{CT} & \texttt{NAIM+ODST} &  68.75 \tiny{± 5.29} & 74.57 \tiny{± 6.42} &  64.42 \tiny{± 4.86} \\
& \texttt{Tabular} & \texttt{RSF}  & 67.47 \tiny{± 3.61} & 75.28 \tiny{± 3.84} & 62.39 \tiny{± 2.20}  \\
& \texttt{Tabular} & \texttt{NAIM+ODST} & $65.52$ \tiny{$\pm$ 4.45} & $72.73$ \tiny{$\pm$ 5.57} & $63.56$ \tiny{$\pm$ 4.35} \\
\midrule[1.5pt]
\multirow{4}{*}{\rotatebox[origin=c]{0}{\scriptsize \texttt{Early}}}
& \texttt{WSI+CT+Tab} & $\scriptstyle^{\bigoplus}\texttt{NAIM+ODST}$ & $73.26$ \tiny{± 4.05} & $77.73$ \tiny{± 5.04} & $68.94$ \tiny{± 5.04} \\
& \texttt{WSI+CT}     & $\scriptstyle^{\bigoplus}\texttt{NAIM+ODST}$ & $71.28$ \tiny{± 4.09} & $74.27$ \tiny{± 5.19} & $67.97$ \tiny{± 4.43} \\
& \texttt{WSI+Tab}    & $\scriptstyle^{\bigoplus}\texttt{NAIM+ODST}$ & $66.65$ \tiny{± 3.08} & $67.77$ \tiny{± 3.15} & $65.00$ \tiny{± 3.60} \\
& \texttt{CT+Tab}     & $\scriptstyle^{\bigoplus}\texttt{NAIM+ODST}$ & $71.93$ \tiny{± 4.74} & \underline{$78.94$ \tiny{± 5.96}} & $68.59$ \tiny{± 3.80} \\
\midrule
% ======================= LATE FUSION =======================
\multirow{4}{*}{\rotatebox[origin=c]{0}{\scriptsize \texttt{Late}}}
& \texttt{WSI+CT+Tab} & \texttt{NAIM+ODST}$^3$ & $71.50$ \tiny{± 4.58} & $76.60$ \tiny{± 5.97} & $67.63$ \tiny{± 3.63} \\
& \texttt{WSI+CT}     & \texttt{NAIM+ODST}$^2$ & \underline{$74.02$ \tiny{± 3.89}} & \underline{$76.52$ \tiny{± 4.60}} & \textbf{70.35 \tiny{± 4.37}} \\
& \texttt{WSI+Tab}    & \texttt{NAIM+ODST}$^2$ & $67.44$ \tiny{± 3.08} & $67.19$ \tiny{± 3.08} & \textbf{70.35 \tiny{± 4.07}} \\
& \texttt{CT+Tab}     & \texttt{NAIM+ODST}$^2$ & $69.00$ \tiny{± 6.11} & $76.93$ \tiny{± 7.80} & $63.86$ \tiny{± 4.80} \\
\midrule
% ======================= INTERMEDIATE FUSION =======================
\multirow{4}{*}{\rotatebox[origin=c]{0}{\scriptsize \texttt{Interm.}}}
& \texttt{WSI+CT+Tab} & \texttt{ConcatODST} & $\mathbf{74.42}$ \textbf{\tiny{± 4.25}} & $\mathbf{80.74}$ \textbf{\tiny{± 5.21}} & \underline{$69.01$ \tiny{± 3.79}}\\
& \texttt{WSI+CT}     & \texttt{ConcatODST} & $73.13$ \tiny{± 3.67} & $77.96$ \tiny{± 4.73} & $68.60$ \tiny{± 3.64} \\
& \texttt{WSI+Tab}    & \texttt{ConcatODST} & $70.63$ \tiny{± 1.41} & $75.34$ \tiny{± 2.64} & $66.77$ \tiny{± 2.56} \\
& \texttt{CT+Tab}     & \texttt{ConcatODST} & $69.90$ \tiny{± 4.94} & $76.32$ \tiny{± 6.11} & $66.09$ \tiny{± 4.51} \\
\bottomrule

\end{tabular}
}
\caption{Survival prediction performance across unimodal and multimodal configurations under different fusion strategies. Models are evaluated using Concordance Index (C-ix), time-dependent AUC (td-AUC), and Uno’s C-index, reported as mean $\pm$ standard deviation over 5-fold cross-validation.
Unimodal baselines include both classical machine learning models and the proposed \texttt{NAIM+ODST} architecture.
Multimodal results are grouped by fusion paradigm: early fusion ($\scriptstyle^{\bigoplus}\texttt{NAIM+ODST}$), late fusion (\texttt{NAIM+ODST}$^n$), and intermediate fusion (\texttt{ConcatODST}).
Best-performing results for each metric are highlighted in \textbf{bold}, while second-best values are \underline{underlined}.}
\label{tab:multimodal_fused_all_metrics}
\end{table*}

%%%%%% ---------------------------------------------------------------------- ----------------------------------------------------------------------
%%%%%% ---------------------------------------------------------------------- ----------------------------------------------------------------------
%%%%%% ---------------------------------------------------------------------- ------------------------
\bl
Building on the unimodal baselines presented in the previous section, Table~\ref{tab:multimodal_fused_all_metrics} summarizes multimodal survival performance under three fusion paradigms, early, late, and intermediate, across all bimodal and trimodal modality combinations. 
A detailed description of all architectures and training procedures is provided in Section~\ref{sec:methods}.

\noindent
Early fusion ($^{\bigoplus}$\texttt{NAIM+ODST}) concatenates the FM-extracted embeddings of all available modalities and feeds the joint vector into \texttt{NAIM+ODST}, the best-performing unimodal DL encoder, trained end-to-end directly on the combined input.
Late fusion (\texttt{NAIM+ODST}$^n$) trains one independent \texttt{NAIM+ODST} per modality and combines their outputs at the decision level by rank-based score aggregation, as described in 
Section~\ref{sec:methods}. 
This ensemble baseline explicitly avoids any feature-level interaction between modalities.
Intermediate fusion (\texttt{ConcatODST}) builds on frozen unimodal encoders pretrained on the same cross-validation fold, with their \texttt{FC} prediction heads removed. 
The latent representations are concatenated and processed by the MM module: a \texttt{FC} layer down-scaling the fused representation by one-third, followed by an \texttt{ODST} layer modeling non-linear cross-modal interactions. 
The freezing strategy and MM module configuration are supported by the ablation study in 
Section \ref{supp:ablation_mm}.
\bb

\bl
The trimodal intermediate fusion model (\texttt{ConcatODST}, \texttt{WSI+CT+Tab}) achieves the highest C-index ($74.42 \pm 4.25$) and td-AUC ($80.74 \pm 5.21$) in this study.
On C-index, it improves by $4.20$ points over the strongest unimodal ML model (CT \texttt{SGB}, $70.22 \pm 5.05$); on td-AUC, by $5.46$ points over the best unimodal result (Tabular \textsc{RSF}, $75.28 \pm 3.84$).
CT features contribute positively across all configurations. 
The \texttt{WSI+CT} late fusion model achieves a C-index of $74.02 \pm 3.89$ and the 
highest Uno C-index among all evaluated models ($70.35 \pm 4.37$), outperforming 
the corresponding early ($71.28 \pm 4.09$, Uno $67.97$) and intermediate 
($73.13 \pm 3.67$, Uno $68.60$) fusion variants, the only configuration 
where late fusion exceeds its intermediate counterparts.
This result is notable given the $64\%$ WSI missingness rate at test time: CT features compensate for absent histopathology, stabilizing predictions that would otherwise depend on the WSI unimodal signal alone (best DL C-index $58.81 \pm 4.04$).

Among bimodal configurations, \texttt{CT+Tab} early fusion reaches a td-AUC of $78.94 \pm 5.96$, the second-highest td-AUC across all models and fusion strategies.
The \texttt{CT+Tab} intermediate model ($69.90 \pm 4.94$) yields a lower C-index than its early counterpart ($71.93 \pm 4.74$), suggesting that joint optimization does not uniformly help when both modalities already carry complementary discriminative signals at the feature level.
By contrast, \texttt{WSI+Tab} underperforms relative to configurations involving CT.
The \texttt{WSI+Tab} intermediate model achieves a C-index of $70.63 \pm 1.41$, lower than both trimodal intermediate ($74.42$) and \texttt{WSI+CT} late ($74.02$), and its td-AUC ($75.34$) falls behind \texttt{CT+Tab} early ($78.94$) by $3.60$ points.
The low variance of the \texttt{WSI+Tab} intermediate configuration (C-index SEM $1.41$) reflects stable but ceiling-limited performance, likely because WSI missingness prevents the histopathological branch from contributing a strong independent signal.

Across fusion strategies, intermediate fusion outperforms early fusion in the trimodal setting ($74.42$ vs $73.26$ in C-index; $80.74$ vs $77.73$ in td-AUC), confirming that end-to-end joint optimization extracts synergies not available to feature-level concatenation.
Late fusion reaches competitive performance only for \texttt{WSI+CT}, where the two unimodal branches are individually strong (CT \texttt{NAIM+ODST} C-index $68.75$; WSI \textsc{RSF} $65.47$); averaging predictions from two well-calibrated models suffices here.
For \texttt{CT+Tab} late fusion, C-index drops to $69.00 \pm 6.11$ despite both unimodal baselines exceeding $65$, confirming that unweighted score averaging is sensitive to mismatched prediction scales across modality types.
\bb

\subsection{Robustness to Modality Dropout and Adaptive Attention}\label{sec-robmissingdrop}

%%%%%% ---------------
\begin{figure*}[!h]

\includegraphics[width=\linewidth]{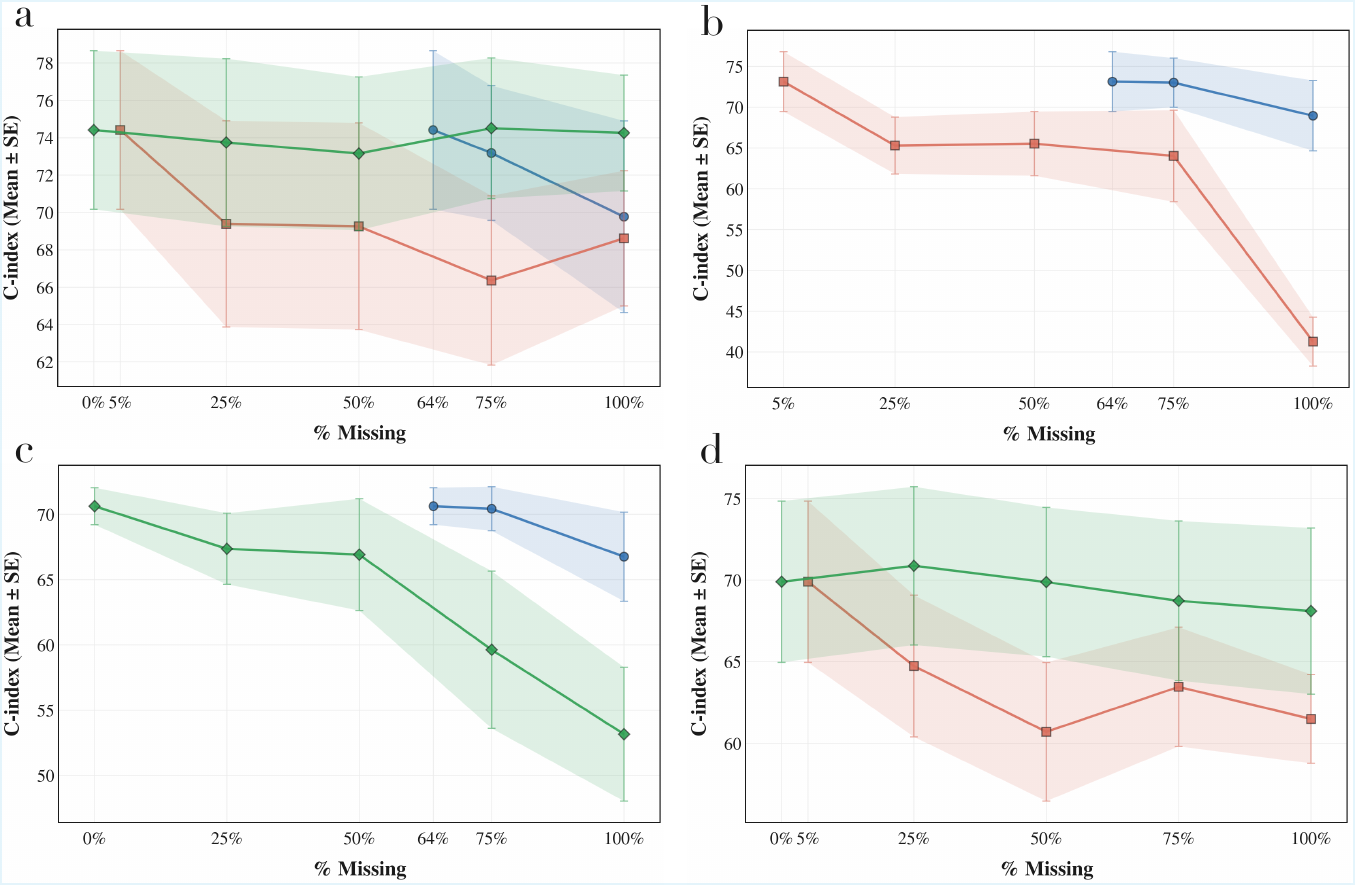}

\caption{Performance of multimodal \textbf{intermediate fusion} models as a function of progressive modality removal.
Each panel reports the mean C-index $\pm$ standard error over 5 folds.
Each line represents the performance when a specific modality is removed at test time, with the x-axis denoting the fraction of test samples for which that modality is missing.
Color coding identifies the modality being removed: green corresponds to Tabular, red to CT, and blue to WSI.
Solid lines denote CT-FM encoders, while dashed lines denote Merlin encoders (panels a, b, d).
Panels correspond to different modality combinations: (a) WSI+CT+Tabular, (b) WSI+CT, (c) WSI+Tabular, and (d) CT+Tabular.}
\label{fig:missingness_multimodal}
\end{figure*}
%%%%%% ---------------

To assess the robustness of the proposed framework under incomplete modality profiles, we investigate how progressive removal of individual data streams affects survival prediction in the trimodal \texttt{ConcatODST} intermediate fusion model.
We conducted a controlled missingness analysis by progressively masking individual modalities in the test set from their natural baseline to complete unavailability ($100\%$ missing), while keeping the remaining modalities at their observed rate.
This evaluation setting serves as a post-hoc Explainable AI (XAI) mechanism. 
Specifically, the performance drop induced by suppressing a modality can be interpreted as an estimate of the model’s reliance on that data stream. 

Figure~\ref{fig:missingness_multimodal}.a decomposes modality contributions in the trimodal setting and reveals a consistent prognostic hierarchy.
Masking Tabular data (green curve) leaves performance stable across the full missingness 
range, suggesting that the WSI and CT streams jointly compensate for absent clinical 
variables in this trimodal setting.
Masking WSI (blue curve) has limited effect until complete removal, where C-index drops 
moderately to $\approx 70$, consistent with the high natural missingness rate already 
present during training.
CT (red curve) is the most sensitive modality: progressive masking drives a consistent 
performance decline reaching a minimum near $75\%$ missing, followed by a partial 
recovery at $100\%$.
Taken together, the trimodal model maintains a C-index above $66$ under complete 
removal of any single modality, demonstrating that the missing-aware fusion design 
yields a model robust to the loss of any individual data stream.

In \textit{WSI+CT} (Fig.~\ref{fig:missingness_multimodal}b), masking WSI causes a 
continuous and sharp degradation reaching $\approx 40$ at complete removal; masking CT 
shows a more moderate decline across the intermediate range before collapsing at 
$100\%$.
Given the substantially higher availability of CT at test time ($95\%$ vs $36\%$ for WSI), 
the model likely relies more heavily on the radiological stream under natural conditions, 
with WSI contributing a complementary signal.
In \textit{WSI+Tabular} (Fig.~\ref{fig:missingness_multimodal}c), masking Tabular data 
drives a steep and monotonic performance drop to $\approx 52$ at $100\%$ missing, while 
masking WSI leaves C-index stable across the full range ($\approx 70$), indicating that clinical variables are the primary driver in this bimodal configuration.
In \textit{CT+Tabular} (Fig.~\ref{fig:missingness_multimodal}d), masking CT produces 
an early performance drop that persists across all missingness levels, whereas masking 
Tabular data has a comparatively small effect, with C-index remaining near the bimodal 
baseline throughout.
Notably, when a bimodal model is forced into a unimodal inference regime (i.e., by masking one modality entirely, performance decreases relative to the full bimodal setting, as expected due to information removal, yet in several cases remains competitive with, or even exceeds, the corresponding unimodal baselines. 
This behavior can be observed in the \textit{WSI+Tabular} where the removal of the WSI slides yields $67.92 \text{±}2,14$ C-index, that is slightly higher than the best performing unimodal ML model trained on Tabular data (\texttt{RSF}, $67.47 \text{±} 3.61$).
This suggests that the model does not merely combine independent unimodal predictors, but learns internal representations that remain informative even when only a subset of modalities is available, supporting the practical utility of the proposed missing-aware fusion approach.

\subsection{Survival Analysis and Disease Progression Stratification}\label{sec-survstrata}
%%%%%% ---------------------------------------------------------------------- ----------------------------------------------------------------------
%%%%%% ---------------------------------------------------------------------- ----------------------------------------------------------------------
%%%%%% ---------------------------------------------------------------------- ----------------------------------------------------------------------
% Stratified KM curves for survival

\begin{figure*}[!h]
 \includegraphics[width=\linewidth]{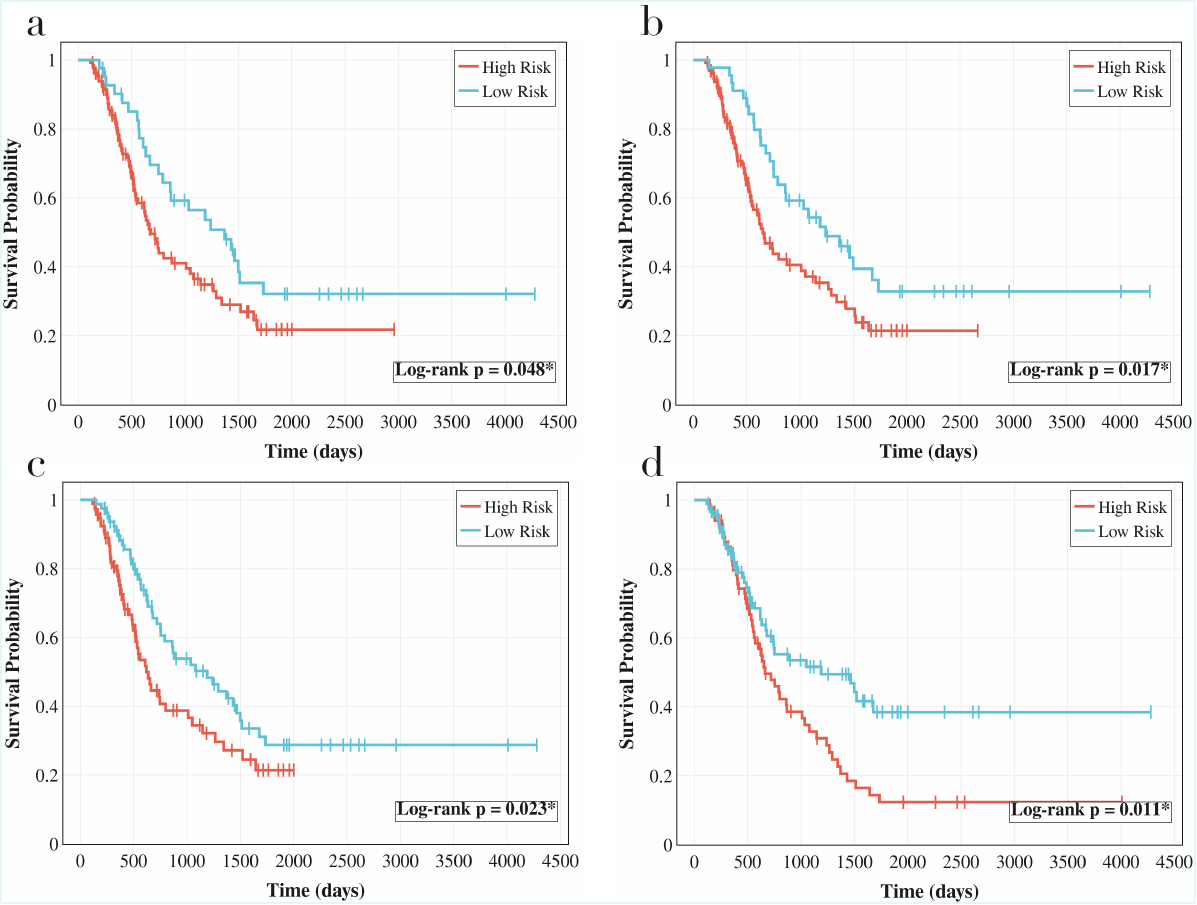}
\caption{Stratified Kaplan–Meier survival curves for intermediate fusion models (\texttt{ConcatODST}), with follow-up time (days) on the x-axis and survival probability on the y-axis.
Patients are stratified into \textit{Low-Risk} and \textit{High-Risk} groups according to the predicted 5-year outcome score produced by each model.
The blue curve represents the \textit{Low-Risk} group, while the red curve corresponds to the \textit{High-Risk} group.
Panels correspond to different modality combinations: (a) CT+Tabular, (b) WSI+CT, (c) WSI+CT+Tabular (trimodal), and (d) WSI+Tabular.}

\label{fig:km_grid}
\end{figure*}

%%%%%% ---------------------------------------------------------------------- ----------------------------------------------------------------------
%%%%%% ---------------------------------------------------------------------- ----------------------------------------------------------------------
%%%%%% ---------------------------------------------------------------------- ----------------------------------------------------------------------

To assess the translational clinical utility of the intermediate fusion models, we performed an aggregated risk stratification analysis.
Risk scores predicted on each validation fold were concatenated into a single distribution, from which an optimal threshold was derived via the log-rank test.
This threshold was then applied to the corresponding test-fold predictions, partitioning patients into \textit{High-Risk} and \textit{Low-Risk} groups.

Kaplan-Meier survival curves were estimated separately for the \textit{High-Risk} and \textit{Low-Risk} groups using the observed survival times, and the log-rank test was applied to assess the statistical significance of the separation between the two strata (Figure~\ref{fig:km_grid}).
All four intermediate fusion configurations achieved significant risk-group discrimination.
CT+Tabular (Figure~\ref{fig:km_grid}.a, $p = 0.048$), WSI+CT 
(Figure~\ref{fig:km_grid}.b, $p = 0.017$), WSI+CT+Tabular 
(Figure~\ref{fig:km_grid}.c, $p = 0.023$), and WSI+Tabular 
(Figure~\ref{fig:km_grid}.d, $p = 0.011$) all achieve statistically 
significant separation between the two risk strata, confirming that 
intermediate fusion produces clinically discriminative risk scores 
across all modality combinations tested.

\begin{figure*}[!t]
    \includegraphics[width=\linewidth]{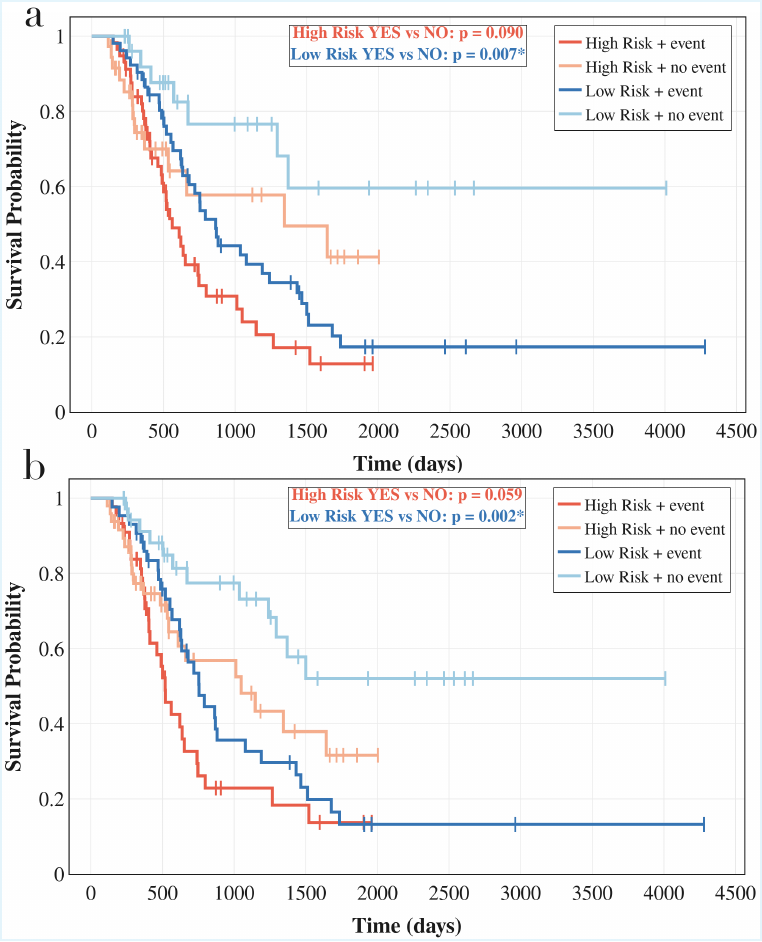}
    \caption{Patients are first stratified into \textit{Low-Risk} and \textit{High-Risk} groups using a threshold derived from validation folds, and this threshold is subsequently applied to the test set.
Within each risk group, patients are further separated according to event status (event vs no-event) for the corresponding endpoint.
This stratification highlights differences in survival trajectories conditioned on both predicted risk and observed progression patterns.
Panel (a) reports progression-free survival (PFS), while panel (b) corresponds to distant metastasis.}
    \label{fig:km_grid_progression}
\end{figure*}

To further validate the clinical relevance of the learned risk scores, we assessed whether the stratification aligns with disease progression patterns beyond overall survival. 
We focused on two complementary endpoints: Progression-Free Survival (PFS), a composite measure capturing local progression, metastatic spread, and censoring status, and Distant Metastasis (DM), which reflects systemic tumor dissemination. 
Figure~\ref{fig:km_grid_progression} reports the Kaplan-Meier curves obtained from the trimodal intermediate fusion model (WSI+CT+Tabular). 
Patient-specific risk scores were aggregated across test folds and used to stratify patients into \textit{Low-Risk} and \textit{High-Risk} groups. Within each risk group, survival differences were evaluated separately for PFS (Figure~\ref{fig:km_grid_progression}.a) and DM (Figure~\ref{fig:km_grid_progression}.b).

For PFS, the Low-Risk group shows a significant separation between patients who experience progression and those who do not (log-rank $p = 0.007$), while in the High-Risk group the separation does not reach significance ($p = 0.090$). 
A similar pattern holds for DM, where the Low-Risk group achieves significant discrimination between patients who develop distant metastases and those who remain metastasis-free ($p = 0.002$), whereas the High-Risk group approaches but does not reach significance ($p = 0.059$).
\bl
The analysis of the resulting survival curves reveals a structural asymmetry across risk strata. 
For both PFS and DM, the no-event subgroups differ sharply between Low-Risk and High-Risk patients: event-free Low-Risk patients sustain high survival probabilities throughout follow-up, while event-free High-Risk patients decline at a substantially faster rate. 
The risk score thus characterizes patients even in the absence of observed progression, capturing latent biological differences that separate indolent from aggressive disease courses. 
By contrast, the event-positive subgroups across both risk strata are largely overlapping, indicating that patients who experience progression or distant metastasis converge toward a common survival trajectory irrespective of their initial risk classification. 

This dual behavior underscores the clinical utility of the proposed 
stratification. 
The model identifies a Low-Risk subpopulation with 
stable disease control that may be managed with reduced surveillance 
intensity. 
In the High-Risk group, the clear separation among 
event-free patients supports the use of the risk score as an early 
warning signal, flagging patients who carry elevated baseline risk 
and may benefit from intensified monitoring or preemptive therapeutic 
escalation before progression manifests.
\bb

\section{Discussion}
\bl

This study addresses a structural gap in multimodal oncology modeling, namely the near-universal assumption of complete modality availability. 
Biopsies are not always feasible, histopathology slides are not always digitized, and CT scans acquired at referring institutions may be unavailable at the treating center. 
Treating missingness as a training design constraint rather than a pre-processing problem allowed us to retain 179 patients in the analysis instead of the 41 who would have remained under a complete-case filter. 
This distinction carries direct clinical consequence. 
The excluded patients are not a random subset. 
Patients with the most aggressive disease trajectories are precisely those for whom full staging workups are most likely to be incomplete, and their systematic exclusion biases any prognostic model toward the more favorable end of the risk spectrum (Sec.~\ref{sec-missing}).

The prognostic hierarchy observed across modalities is consistent with known biological properties of locally advanced NSCLC. 
Structured clinical variables, which encode treatment decisions, molecular driver status, and performance status, carry prognostic information that imaging alone cannot replicate (Section~\ref{cohort}). 
The significant univariate associations of EGFR mutation status ($p = 0.033$), radiotherapy technique ($p < 0.001$), and adjuvant immunotherapy ($p = 0.010$) with 5-year survival reflect the extent to which treatment-era factors, rather than baseline tumor biology alone, determine outcome in this population. 
The survival advantage of modern volumetric arc techniques over three-dimensional conformal radiotherapy and the impact of consolidation durvalumab following the PACIFIC regimen are well-established clinically; their reflection in the tabular stream confirms that the model is not learning noise. 

From a clinical perspective, the most important question raised by this work is how a continuous risk score can be made actionable within existing oncology workflows. 
The stratification analysis showed significant separation between High-Risk and Low-Risk groups across all four intermediate fusion configurations (log-rank $p$ ranging from 0.011 to 0.048), providing statistical evidence of discriminative capacity, but statistical significance alone does not determine clinical utility. 
Oncologists managing unresectable Stage III NSCLC operate within a relatively narrow therapeutic decision space.
The clinical value of a risk score in this setting therefore lies not in treatment selection per se, but in surveillance intensity, early escalation triggers, and patient counseling. 
The within-group PFS and DM stratification produced by the trimodal model enables this pathway more concretely. 
Within the Low-Risk group, patients who subsequently progressed separated from those who remained event-free with high significance ($p = 0.007$ for PFS, $p = 0.002$ for DM), while the High-Risk group showed a faster overall decline even among patients who did not progress (Sec.~\ref{sec-survstrata}). 
This asymmetry implies that the risk score is not simply summarizing binary outcome status, but captures a latent biological aggressiveness that manifests as trajectory rather than event. 
Clinically, a High-Risk, event-free patient is not the same as a Low-Risk, event-free patient, even when their observed histories are identical at the time of scoring. 
This distinction is precisely the kind of information that could justify earlier post-treatment imaging in High-Risk patients, or enrollment into consolidation therapy trials for those with elevated baseline risk scores who nonetheless complete chemo-radiotherapy without acute progression.

Translating these risk scores into clinical practice requires threshold definitions that go beyond the log-rank-optimized binary cutoff used in this retrospective analysis. 
The optimal cutoff derived from validation-fold score distributions served as a methodological tool for assessing stratification capacity, not as a clinically calibrated decision boundary. 
For prospective use, thresholds should be defined relative to institutional event rates and treatment protocols. 
Risk score integration would be most useful as a continuous variable overlaid on existing stratification systems such as TNM staging and ECOG performance status, not as a replacement for either. 
The practical deployment pathway We envision is one where the multimodal risk estimate is surfaced within the tumor board workflow, alongside the radiological read and pathological report, as an additional quantitative input that extends the prognostic horizon without displacing established clinical reasoning. 
Importantly, the model's tolerance for missing modalities means it can produce a score for any patient who has at least tabular data available, which is the universal modality in this cohort. 
This property is essential for clinical adoption, as a tool that silently fails or refuses inference when a slide is unavailable is not usable in routine care.

Several open questions condition whether the performance reported here translates beyond this cohort. 
The study population is deliberately narrow, comprising unresectable Stage II-III NSCLC treated with radical chemo-radiotherapy at a single institution. 
The exclusion criteria that define this population produce a clinically homogeneous group but limit generalization. 
The prognostic dynamics of resectable Stage II disease, where surgery is the primary treatment, or of metastatically Stage IV disease managed with SBRT and systemic therapy, differ substantially from those modeled here. 
Multi-center validation within the same clinical indication is the prerequisite for generalizability, and it requires partner institutions that share not only tumor boards and treatment protocols but also data infrastructure that permits harmonization of CT acquisition parameters, WSI digitization standards, and tabular variable definitions across sites.

Beyond cohort scope, two technical directions represent the most consequential next steps. 
The first is tumor-aware CT feature extraction. 
All three foundation models evaluated in this study encode the full thoracic volume, treating gross tumor burden, mediastinal involvement, and uninvolved lung parenchyma as an undifferentiated signal (Section~\ref{supp:ct_fms}).
Restricting extraction to the gross tumor volume or clinical target volume, as defined by the treating radiation oncologist, would concentrate the radiological representation on the tissue directly responsible for local failure and metastatic seeding. 
This is not merely an engineering refinement, given that GTV delineation already exists in the treatment planning workflow for every patient who undergoes radical chemo-radiotherapy, making it a zero-cost additional input from a data collection standpoint. 
Whether foundation models conditioned on anatomically restricted inputs recover prognostic signal that whole-volume encoders suppress is an empirical question this study cannot answer, but the biological rationale for testing it is strong. 
The second direction concerns fusion expressiveness. 
The intermediate fusion architecture used here, a concatenation-based head with frozen unimodal encoders, was chosen for its interpretability and resistance to overfitting on a small cohort. 
Cross-attention mechanisms that allow modality representations to condition one another during feature refinement, or dynamic gating networks that weight each stream by its observed completeness and estimated informativeness per patient, would in principle extract richer cross-modal interactions. 
These approaches require larger missing-aware multimodal cohorts to train reliably without collapsing to degenerate solutions, and the field does not yet have well-validated benchmarks for this setting. 
A clear priority is therefore the construction of multi-institutional datasets with heterogeneous modality profiles collected under standard clinical workflows.
Computational feasibility is a practical barrier to clinical translation that has received less attention than model architecture. 
Foundation model inference for a single patient in this pipeline requires WSI tissue segmentation, patch-level encoding across thousands of image tiles, slide-level aggregation, and volumetric CT embedding, none of which is compatible with real-time clinical use under current hardware constraints. Knowledge distillation of the imaging FMs into lightweight surrogates, or caching of embeddings computed at the time of treatment planning, would reduce inference latency to a degree compatible with tumor board timelines. The tabular and survival head components are computationally trivial; the bottleneck is entirely in the FM extraction stage.

Finally, the incremental value of the multimodal risk score over established prognostic indices remains to be quantified. 
This study benchmarks modality combinations against one another and against unimodal baselines, but does not include direct comparison against \textit{TNM} stage, \textit{ECOG} performance status, or published nomograms for NSCLC~\cite{zhao2024clinical}. 
Demonstrating net reclassification improvement over these existing tools, or equivalently showing that the multimodal score stratifies patients within TNM substages where clinical staging provides insufficient discrimination, is the evidence threshold that would move this class of model from methodological interest to clinical relevance. 
These comparisons are the natural next step once a prospectively collected, multi-site cohort with complete staging and outcome data is available.

\bb

\section{Methods}\label{sec:methods}

\subsection{Materials and Cohort Selection}
Data collection was carried out at the \textit{Fondazione Policlinico Universitario Campus Bio-Medico of Rome} and encompassed 353 patients with histologically confirmed Non–Small Cell Lung Cancer (NSCLC), all diagnosed at stage~II–III following the 9th edition TNM classification of lung cancer~\cite{detterbeck2024proposed} and enrolled retrospectively or prospectively between 2012 and 2024. 
From this cohort, we retained only individuals classified as unresectable stage~II–III disease~\cite{Daly2021Management} (IIA, IIB, IIIA, IIIB, or IIIC) using the 9th edition of the TNM classification for Thoracic Oncology~\cite{detterbeck2024proposed}. 
Accordingly, the study employed the following exclusion criteria to ensure a homogeneous cohort: (i) patients with resectable tumors who underwent surgery in neoadjuvant settings; (ii) patients treated with palliative intent; (iii) cases with incomplete baseline clinical information or missing follow-up; (iv) small cell lung cancers patients.
In addition, we removed patients with non-NSCLC histologies, metastatic or recurrent disease at presentation, prior thoracic surgery or postoperative RT/RTCT, those managed with non-standard therapeutic pathways (e.g., SBRT alone or systemic therapy only), and cases lacking verifiable diagnostic imaging or pathology.

After applying these criteria, the final study population consisted of 179 patients, providing a homogeneous cohort representative of the clinical course of locally advanced, non-resectable NSCLC.
For every included patient, a comprehensive set of structured clinical variables was curated to capture the biological profile and treatment history. 
As detailed in Tables~\ref{tab:baseline_characteristics_part1} and~\ref{tab:baseline_characteristics_part2}, data collection was stratified into four key domains: (i) \textit{general characteristics}, including demographic and physiological baselines such as sex, age, weight, height, smoking status, daily cigarette consumption, family history of neoplasms, numeric pain rating scale (NRS), and specific comorbidity profiles (vascular, metabolic, renal, pulmonary); (ii) \textit{tumor biology}, encompassing histological subtype (adenocarcinoma, squamous cell carcinoma, NOS), TNM classification, molecular biomarkers (EGFR, ALK, MET, PD-L1), and other biopsy details; (iii) \textit{treatment parameters}, covering chemotherapy details (induction cycles, specific drug schemes), radiotherapy metrics (total dose, fractionation, technique, duration), treatment interruptions, and the administration of adjuvant immunotherapy; and (iv) \textit{toxicity and outcomes}, recording grade-specific adverse events (esophageal, pulmonary, hematological).

Beyond Overall Survival (OS), the study incorporated a comprehensive assessment of disease progression patterns to characterize failure modes. 
Secondary clinical endpoints included: (i) \textit{distant Metastasis (M)}, indicating systemic disease spread to non-contiguous organs; and (ii) \textit{Progression-Free Survival (PFS)}, calculated as the time from treatment initiation to the first documented event of local and distant progression or death.

\noindent
\textbf{Imaging Acquisition.}
For each patient, multimodal imaging data were systematically acquired prior to the initiation of radical therapy to establish a baseline for prognostic modeling.
\begin{itemize}
    \item \textbf{Radiology.} 
    CT scans were retrospectively retrieved from the institutional Picture Archiving and Communication System (PACS). 
    Specifically, we collected the \textit{treatment planning CTs} acquired within 30 days prior to the start of Chemoradiotherapy. 
    Unlike standard diagnostic scans, these planning CTs are acquired under rigorous protocols to ensure geometric accuracy for dose calculation, providing a standardized and reproducible representation of the macroscopic tumor burden and thoracic anatomy across the entire cohort.
    \item \textbf{Pathology.} 
    High-resolution Whole Slide Images (WSIs) were generated from diagnostic tissue biopsies obtained via bronchoscopy or CT-guided percutaneous needle aspiration. 
    Formalin-fixed paraffin-embedded (FFPE) tissue blocks were sectioned and stained with Hematoxylin and Eosin (H\&E) according to standard clinical workflows. 
    Subsequently, the physical glass slides were digitized at $20\times$ magnification ($0.5\,\mu$m/pixel) using a high-throughput slide scanner. 
    This digitization process transformed the analog histological samples into gigapixel-resolution digital slides, ensuring the preservation of the intricate microscopic tumor architecture and cellular phenotypes required for computational analysis.
\end{itemize}
\subsection{Ethical approval}
This study was conducted in accordance with the Declaration of Helsinki. 
Ethical approval was obtained from the Ethical Committee of \textit{Fondazione Policlinico Universitario Campus Bio-Medico}.
Data collection spans two regulatory phases: a retrospective phase approved on 30 October 2012 and registered on \textit{ClinicalTrials.gov} (Identifier: NCT03583723), and a prospective phase approved on 16 April 2019 (Identifier: 16/19 OSS).
All patients provided written informed consent for the use of their clinical and imaging data for research purposes.

\subsection{Methodological Framework Overview}

We propose a three-stage multimodal framework designed to predict survival outcomes from incomplete heterogeneous data. 
As illustrated in Figure~\ref{fig:method_architecture}, the pipeline consists of: (1) Unimodal feature extraction using pre-trained FM to bypass data scarcity; (2) Missing-aware representation learning using the \texttt{NAIM} architecture; and (3) our proposed intermediate fusion strategy.

\label{sec:pipeline}

\subsection{Step 1: Preprocessing \& Feature Extraction}

We formalize the three-stage multimodal learning pipeline illustrated in Fig.~\ref{fig:method_architecture}. Given three heterogeneous input modalities, Tabular (\(x_1\)); CT (\(x_2\)), and WSI (\(x_3\)), the goal is to construct a unified representation for downstream survival prediction.
We denote by \(\Phi_i : \mathcal{X}_i \rightarrow \mathbb{R}^{d_i}\) the modality-specific feature extractors producing high-level embeddings \(x_i^\ast \in \mathbb{R}^{d_i}\) leveraging state of the art FMs for each modalities or classical pre-processing pipelines. 

\noindent
\textbf{Tabular Data (\(x_1\)).}
To ensure consistency during model training, all clinical features underwent a standardized preprocessing pipeline. 
Categorical variables (e.g., comorbidities, histology) were one-hot encoded, while ordinal features (e.g., disease stage, ECOG status) were integer-encoded to preserve their inherent hierarchy. 
Numerical features (e.g., age, weight) were standardized via z-score normalization to zero mean and unit variance. The resulting feature vector is denoted as \(x_1^\ast = \Phi_1(x_1)\), where \(\Phi_1\) represents the composite preprocessing function.

\textbf{Radiology Imaging (\(x_2\)).}
Volumetric radiological features are extracted using CT-FM~\cite{pai2025vision}, a SegResEncoder-based FM pretrained via intra-sample contrastive learning (SimCLR) on 148{,}000 CT scans from the Imaging Data Commons. 
Raw CT volumes undergo the following preprocessing steps, consistent with the CT-FM encoding pipeline:
(i)~\textit{Orientation:} all volumes are reoriented to a canonical SPL (Superior-Posterior-Left) coordinate system.
(ii)~\textit{Windowing \& Normalization:} voxel intensities are clipped to the range $[-1024, 2048]$\,HU and linearly scaled to the $[0, 1]$ interval.
(iii)~\textit{Foreground cropping:} background regions are removed via foreground detection, retaining only the anatomically informative portion of the volume without imposing a fixed spatial resolution or crop size.
The preprocessed volume \(x_2\) is passed through the CT-FM encoder \(\mathcal{E}_{CT}\), and the resulting feature maps are reduced to a single embedding via global average pooling, yielding a $512$-dimensional representation:
\begin{equation}
x_2^\ast = \Phi_2(x_2) = \mathrm{AvgPool}\!\bigl(\mathcal{E}_{CT}(x_2)\bigr)
\in \mathbb{R}^{512}
\end{equation}

\noindent
\textbf{Pathology Imaging (\(x_3\)).}
To capture the microscopic tumor characterization, we utilize a hierarchical encoding pipeline to extract a unique representation vector for each WSI. 
First, WSIs are processed using CLAM~\cite{CLAM} (CLustering-constrained Attention Multiple instance learning) with the objective to segment tissue from background artifacts. 
From each WSI ($x_3$), we extract a slide-specific set of tissue-containing image tiles. After tissue segmentation, all background and artefactual regions are removed, and the remaining tissue area is decomposed into a non-uniform number $N_w$ of non-overlapping patches of size $256 \times 256$ pixels at $20\times$ magnification. 
For each $x_3$, we denote this patch set as $P_w (x_3) = \{ p_j \}_{j=1}^{N_w}$, where each element $p_j$ corresponds to a spatially localized region that retains diagnostically meaningful microscopic information.
This patch-based decomposition allows the model to focus on localized morphological patterns that would otherwise be diluted at whole-slide scale.

Each patch $p_j \in P_w$ is then projected into a descriptive 512-dimensions embedding using the vision encoder of MI-Zero~\cite{conch} (\(\mathcal{E}_{patch}\)), a vision-language foundation model pre-trained to align visual features with semantic pathology descriptions. 
Eq.~\ref{eq_patch} formalizes the projection step:
\begin{equation}
    \mathbf{z}_j = \mathrm{\mathcal{E}_{patch}}(p_j) \in \mathbb{R}^{512}, \qquad p_j \in P_w
\label{eq_patch}
\end{equation}

Finally, TITAN~\cite{titan}(Transformer-based pathology Image and Text Alignment Network), \(\mathcal{E}_{wsi}\), aggregates the sequence of patch embeddings into a unified latent representation with $768$ dimensions. 
TITAN leverages a self-supervised masked prediction task to coherently synthesize local cellular phenotypes into a global descriptor. 
We indicate the final slide-level feature vector in the following Eq.\ref{eq_wsi}:
\begin{equation}
x_3^\ast = \mathcal{E}_{wsi}\big(\{\mathbf{z}_j\}_{j=1}^{N_{w}}\big) \in \mathbb{R}^{768}.
\label{eq_wsi}
\end{equation}
By structuring the WSI into interpretable units and encoding them consistently, this pipeline preserves fine-grained microscopic details while ensuring that the representation is compatible with the subsequent multimodal fusion stages. 
The complete preprocessing pipeline $\Phi_3$, representing the composition of patching, encoding, and aggregation, is formalized as:
\begin{equation}
x_3^\ast = \Phi_3(x_3) = \mathcal{E}_{\text{wsi}}\left( \mathcal{E}_{\text{patch}}(p_j) \right) \in \mathbb{R}^{768}; {p_j \in P_w (x_3)} .
\end{equation}

\subsection{Step 2: Missing-Aware Unimodal Encoding (\texttt{NAIM})}
\label{sec:NAIM}

To robustly accommodate structural missingness, we employ the Not Another Imputation Method (\texttt{NAIM})~\cite{NAIM} model.
Standard deep learning models require complete input vectors; traditionally, this forces researchers to use \textit{imputation} (replacing missing values with means, medians, or synthetic guesses), which inevitably introduces noise and bias.
\texttt{NAIM} offers a fundamental paradigm shift: rather than synthesizing fake data, the network is architected to \textit{dynamically ignore} absent information. In our framework, we instantiate a separate \texttt{NAIM} encoder \(F_i\) for each modality. Each encoder takes as input the raw feature vector \(x_i^\ast\) (from Step 1) and a binary mask \(M_i \in \{0,1\}^K\) (where 1 indicates missing), projecting them into a robust latent representation \(h_i\).

\noindent
\textbf{Dual-Stream Feature Embedding.}
The first challenge is to convert heterogeneous data (continuous numbers, categorical classes, and missing values) into a unified high-dimensional space where they can interact. 
\texttt{NAIM} achieves this via a dual embedding strategy that treats missingness not as a null value, but as a specific structural state.

\begin{itemize}
    \item \textbf{Categorical Features (\(x^{cat}\)):}
    Discrete variables (e.g., histology, sex) are mapped using a learnable lookup table \(E^{cat}\). If a feature value is present, the lookup table selects the corresponding trainable embedding vector; if the value is missing, the encoder instead retrieves the frozen padding vector. 
    Formally, for the \(j\)-th categorical feature, the embedding is defined in Eq.~\ref{eq:cat_emb}.
    \begin{equation}
    \label{eq:cat_emb}
    e_j^{cat} = b_j + E_j^{cat}(x_j^{cat}),
    \end{equation}
    where \(b_j\) is a learnable bias term specific to that feature index.

    \item \textbf{Numerical Features (\(x^{num}\)):}
    Continuous variables (e.g., age, tumor volume) require a more nuanced approach to preserve their magnitude. 
    We employ a specialized lookup table \(E^{num}\) containing only two vectors: \(\mathbf{v}_{present}\) (trainable) and \(\mathbf{v}_{missing}\) (frozen zeros).If the feature is missing, the embedding is simply the bias term plus the zero vector.
    If the feature is present, we multiply the trainable \(\mathbf{v}_{present}\) vector by the scalar value of the feature \(x_j^{num}\).
    We formulate the numerical embedding in the following equation:
    \begin{equation}
    \label{eq:num_emb}
    e_j^{num} = b_j + x_j^{num} \cdot E_j^{num}(\mathbb{I}_{present}),
    \end{equation}
    where \(\mathbb{I}_{present}\) is the indicator index. 
    This mechanism allows the network to learn the \textit{significance} of the feature via \(\mathbf{v}_{present}\) while scaling it by the actual patient data.
\end{itemize}

\noindent
\textbf{Double-Sided Masked Attention.}
Once the data is tokenized into embeddings, it is processed by the \texttt{NAIM} Transformer encoder.
However, standard self-attention mechanisms would fail because missing tokens would still participate in the softmax normalization, corrupting the attention distribution.
\texttt{NAIM} solves this via a double-sided masking strategy. 
The goal is to ensure that a missing feature neither \textit{casts} attention (influencing others) nor \textit{receives} attention (being updated by others).

Let \(Q, K, V\) be the query, key, and value matrices derived from the input embeddings. 
We define a mask matrix \(\mathbf{M}\), where \(\mathbf{M}_{ij} = -\infty\) if either feature \(i\) or feature \(j\) is missing, and \(0\) otherwise. 
The resulting formulation, reported in Eq.~\ref{eq:naim_attention}, applies the mask before and after the softmax operation, thereby enforcing a complete
suppression of missing-feature interactions:

\begin{equation}
\label{eq:naim_attention}
\mathrm{A}(Q,K,V) = 
\mathrm{ReLU}\!\left(
\mathcal{S}\!\left(
\frac{Q K^\top}{\sqrt{d_h}} + \mathbf{M}
\right) + \mathbf{M}^\top
\right)V.
\end{equation}

Where $\mathcal{S}$ is the softmax operator. 
This bidirectional exclusion guarantees that the attention mechanism operates solely on observed features, preserving the functionality of
the learned representations even under substantial missingness.
Here is how this mechanism works step-by-step:
\begin{enumerate}
    \item \(\frac{QK^T}{\sqrt{d_h}} + \mathbf{M}\): Adding \(-\infty\) to the attention scores forces the corresponding softmax, values to exactly zero. 
    \item \(+ \mathbf{M}^\top\): we add the mask again (transposed) after the softmax. This is a crucial reinforcement step that explicitly zeros out any residual connection \textit{from} a missing feature to others.
    \item \(\mathrm{ReLU}\): Ensures non-negativity and sparsity in the final attention map.
\end{enumerate}
This mechanism ensures that gradients propagate \textit{only} through observed data paths, effectively creating an architecture that changes structure based on which data are available for a specific patient.

\noindent
\textbf{\texttt{NAIM} Encoders for Unimodal Survival Prediction}
Each modality is processed independently using a dedicated \texttt{NAIM} encoder. 
Unlike standard architectures that treat the input as a monolithic vector, \texttt{NAIM} tokenizes the input modality \(x_i\) into a sequence of distinct feature embeddings (as defined in Eqs.~\ref{eq:cat_emb}--\ref{eq:num_emb}). 
These tokens are processed via the masked self-attention mechanism, which captures intra-modality correlations while strictly preventing information leakage from missing values.
The resulting sequence of contextualized embeddings is then flattened to form the comprehensive unimodal representation \(h_i \in \mathbb{R}^{d_{model}}\).

To translate the robust latent representation \(h_i\) into a continuous survival prediction, the embedding is passed through a hybrid head composed of an ODST~\cite{popov2019neural}, noted as \(E_i \), followed by a Fully Connected (FC) layer.
The ODST architecture bridges the gap between the interpretability of decision trees and the gradient-based optimization of DL. 
Unlike standard decision trees, which rely on greedy, hierarchical splits, an \textit{oblivious} tree applies a uniform splitting criterion across all nodes at a given depth.
In this differentiable implementation, this hard splitting logic is relaxed using a soft gating function (e.g., sigmoid), enabling the entire ensemble to be optimized end-to-end via backpropagation.
Formally, the ODST block maps the unimodal embedding \(h_i\) to a structured intermediate representation, capturing non-linear interactions between features. 
This transformed vector is then projected by a the FC layer to produce the scalar risk score \(y_i\), as defined in Eq.~\ref{eq:prediction_head}:
\begin{equation}
\label{eq:prediction_head}
y_i = \mathrm{FC}\Big(\mathrm{E_i}(h_i)\Big).
\end{equation}

\noindent
\textbf{Optimization Objective.}
Based on the Cox proportional hazards model, The entire pipeline is trained end-to-end by minimizing the negative log partial likelihood loss~\cite{tong2020deep}.
Let \(N_{ob}\) be the total number of observed events in a batch. 
The loss function \(\mathcal{L}_{sur}\) is defined as:

\begin{equation}
\label{eq:cox_loss}
\mathcal{L}_{sur} = - \frac{1}{N_{ob}} \sum_{i : C_i = 1} \left( y_i - \log \sum_{j : T_j \ge T_i} e^{y_j} \right),
\end{equation}

where:
\begin{itemize}
    \item \(C_i = 1\) indicates the occurrence of the event for patient \(i\) (uncensored).
    \item \(y_i\) is the hazard risk score predicted by the network for patient \(i\).
    \item \(T_i\) and \(T_j\) denote the observed survival times for patient \(i\) and \(j\), respectively.
    \item The inner summation \(\sum_{j : T_j \ge T_i}\) aggregates the risk over the set of all patients \(j\) still at risk at time \(T_i\).
\end{itemize}

This objective enforces the correct relative ordering of survival times by penalizing the model when a patient who experienced an event earlier is assigned a lower risk score than those who survived longer. 
\subsection{Step 3: Intermediate Fusion and Multimodal Survival Prediction}
The framework subsequently transitions to the proposed \textbf{intermediate fusion} strategy.
Each modality is processed by its dedicated \texttt{NAIM} encoder and ODST layer (as detailed in Sec.~\ref{sec:NAIM}), producing three latent representations: $h_1$ (Tabular), $h_2$ (CT), and $h_3$ (WSI). 
These embeddings capture modality-specific abstractions while being explicitly conditioned to handle missingness via the double-sided masking mechanism (Eq.~\ref{eq:naim_attention}). 
The unimodal encoders are loaded from Step~2 with their weights frozen throughout multimodal training, preserving the modality-specific representations learned during unimodal pretraining and reducing 
the risk of overfitting on the limited cohort.
To construct the unified multimodal patient profile, the frozen latent vectors are fused via concatenation ($\oplus$):
\begin{equation}
\label{eq:multimodal_fusion}
h_m = \oplus(\, h_1 ,\,\, h_2,\, h_3 \,) \in \mathbb{R}^{d_1+d_2+d_3}.
\end{equation}
This fused vector $h_m$ retains complementary prognostic signals from each data source. 
In the proposed multimodal configuration, referred to as \texttt{ConcatODST}, $h_m$ is passed to the multimodal module $E_m$, which is the only trainable component in this stage and maps the concatenated representation to the final risk score $y_m$. 
$E_m$ comprises two sequential layers $L_1$ and $L_2$ followed by a third fully connected layer ($L_3$), mirroring the unimodal head structure. 
$L_1$ is a FC bottleneck that projects $h_m$ to one-third of its input dimensionality, while $L_2$ is an ODST block that captures non-linear interactions across the fused unimodal embeddings. 
The full ablation over $E_m$ architecture choices is reported in Section~\ref{supp:ablation_mm}. 
The $E_m$ mathematical formulation follows:

\begin{equation}
y_m = \mathrm{FC}\ \!\Big(\,L_2\ \!\Big(\,L_1\ \!\big(\,h_m\,\big)\Big)\Big) = E_m(h_m)
\end{equation}
$E_m$ is optimized end-to-end using the $\mathcal{L}_{sur}$ objective defined in Eq.~\ref{eq:cox_loss}, with gradients propagating exclusively through the multimodal head while the upstream unimodal encoders remain fixed.

\subsection{Model Comparison and Experimental Setup}
To evaluate the proposed intermediate fusion framework, we conducted a structured benchmarking analysis comparing different architectural paradigms, integration strategies, and established ML baselines.

\noindent
\textbf{Unimodal Baselines and Ablation Study.}
Before assessing multimodal fusion, we examined the performance of each modality independently through a unimodal ablation study. 
For Tabular, CT, and WSI data, we compared the following configurations:
\begin{itemize}
    \item Machine learning baselines: Cox Proportional Hazards (\texttt{CPH}), Random Survival Forests (RSF), and \texttt{SGB}, trained on the extracted features.
    \item Standard Neural Network (\texttt{NN}): a baseline model using the feature embedding layer, as described in Section~\ref{sec:NAIM}, followed by a multi-layer perceptron with ReLU activations.
    \item \texttt{NAIM} (encoder-only): the \texttt{NAIM} encoder paired with a linear survival head, isolating the contribution of the masked attention mechanism.
    \item \texttt{NAIM+ODST}: the complete unimodal model combining the \texttt{NAIM} encoder with the \texttt{ODST} survival head.
\end{itemize}

\noindent
\textbf{Multimodal Fusion Strategies.}
To isolate the contribution of the intermediate fusion architecture, we benchmarked it against two alternative integration paradigms, adopting a consistent integration framework (concatenation + ODST head) but varying the optimization scope:
\begin{itemize}
 \bl
    \item \textbf{Early Fusion ($\scriptstyle^{\bigoplus}\texttt{NAIM+ODST})$} Integration is performed directly at the feature level. 
    The FM-extracted embeddings from Step~1 are concatenated into a single unified vector, $x_{\text{early}} = \oplus( x^*_1, x^*_2, x^*_3 )$, which is fed into a single \texttt{NAIM+ODST} encoder trained end-to-end on the combined input. 
    This baseline tests whether a single missing-aware encoder trained jointly on the concatenated feature space can match the performance of architectures that process each modality independently before fusion.\bb

    \item \textbf{Intermediate Fusion (ConcatODST):} In this proposed approach, the architecture is identical to early fusion, but optimization is performed \textit{end-to-end}. 
    The unimodal encoders remain trainable, allowing the backpropagation of gradients from the multimodal ODST head through to the modality-specific branches. 
    This enables the encoders to adapt their feature representations to maximize cross-modal synergy.

    \item \textbf{Late Fusion ($\texttt{NAIM+ODST}^{n}$):} Integration at the prediction level. Independent unimodal models (\texttt{NAIM+ODST}) are trained separately for each available modality. 
    At inference time, we employ a \textit{ranking-based ensemble}: risk scores from each unimodal model are converted into ranks (ordering the cohort from lowest to highest risk). The final multimodal risk score is the sum of these ranks:
    \[
    \text{Rank}_{\text{final}} = \sum_{m \in \mathcal{M}} \text{Rank}(y_m),
    \]
    where \(\mathcal{M}\) is the set of available modalities. This non-parametric aggregation assumes no synergistic interaction between modalities.
\end{itemize}

\noindent
\textbf{Experimental Setup.}
To mitigate overfitting and ensure robust performance estimation, we employed a stratified 5-fold cross-validation strategy. 
The dataset was partitioned into 5 non-overlapping folds, preserving the ratio of events to censored patients. 
In each iteration, 3 folds were used for training, 1 for validation and 1 for testing.
This procedure was repeated 5 times, and final metrics are reported as the mean $\pm$ standard deviation.
Optimization utilized the \textit{AdamW} optimizer (weight decay $1 \times 10^{-5}$) with a batch size of 32 for all the training approaches.
Training ran for a maximum of 500 epochs with early stopping (patience equal to 50 epochs) based on the loss computed on the validation set.
To ensure a fair comparison across fusion paradigms, we adopted a unified two-stage training protocol. In the early fusion setting (ConcatODST\textsuperscript{$\star$}), the \texttt{NAIM} encoders were initialized using weights obtained during unimodal pre-training and kept frozen throughout training. In the intermediate fusion setting (ConcatODST), the encoders were initialized identically but remained trainable, enabling end-to-end optimization in which gradients from the survival loss refine modality-specific representations.

The learning rate schedule ($\eta$) was defined within the range $\eta_{\min}=10^{-8}$ to $\eta_{\max}=10^{-5}$ and consisted of three components. 
First, a linear warm-up over the initial 50 epochs increased $\eta$ from $\eta_{\min}$ to $\eta_{\max}$ to avoid early training instability. 
After warm-up, a plateau detector monitored the validation loss; if no improvement was observed for 20 epochs, a decay schedule was initiated to anneal $\eta$ back to $\eta_{\min}$ across 12 steps. 
Training was halted if the validation metric failed to improve for 50 consecutive epochs.

During end-to-end optimization of the intermediate fusion model, we adopted differential learning rates to balance stability and plasticity. 
The multimodal ODST and fully connected layers were updated using the full learning-rate range (up to $\eta_{\max}=10^{-5}$), while the pretrained unimodal encoders used a tenfold smaller rate (effective $\eta_{\max}=10^{-6}$), preserving established feature representations while allowing controlled adaptation during cross-modal alignment.
All models were implemented in PyTorch and trained on a single \textit{NVIDIA A40 GPU} (48GB VRAM). 

\noindent
\textbf{Evaluation Metrics.}
Model discrimination was quantified using the \textit{Concordance Index (C-index)}, which measures the probability that, for a random pair of subjects, the subject who experienced the event earlier has a higher predicted risk score. 
To rigorously account for potential censoring bias, we reported also the \textit{Uno's C-index}~\cite{Park2021Review}. 
The latter relies on \textit{Inverse Probability of Censoring Weights} (IPCW) to correct for the unequal probability that individuals remain uncensored over
time. 
By weighting each observation by the inverse of its estimated probability of being uncensored, IPCW compensates for informative censoring and yields a consistent measure of discriminative performance even when the censoring mechanism depends on patient covariates.
Lastly, to assess predictive performance over the follow-up period, we computed the \textit{time-dependent Area Under the Curve} (td-AUC). 
Specifically, we evaluated the mean cumulative dynamic AUC at the 25th, 50th, and 75th percentiles of the observed event times in the test set. 

\noindent
\textbf{Statistical Analysis.}
Continuous variables are reported as mean $\pm$ standard deviation, categorical variables as frequencies ($n$, \%). Baseline comparisons utilized the Chi-square test (categorical) and one-way ANOVA (continuous).
Survival curves were estimated via the Kaplan-Meier method and compared using the Log-Rank test.
Statistical significance was set at a two-sided $p < 0.05$. 
All analyses were conducted in Python (sciPy, lifelines and scikit-learn).

%% --- BEGIN DECLARATIONS BLOCK ---
 
\section*{Data Availability}
The dataset analyzed in this study was collected and used under institutional ethical approval and informed patient consent. 
The data are available upon reasonable request for non-commercial research purposes only, addressed to the corresponding authors. 
 
\section*{Code Availability}
The code implementing the proposed framework will be made publicly available upon publication at \url{https://github.com/fruffini/Handling_Missing_Modalities_NSCLC}. 
 
\section*{Acknowledgments}
Filippo Ruffini is a PhD student enrolled in the National PhD in Artificial Intelligence, XXXVIII cycle, course on Health and life sciences, organized by Università Campus Bio-Medico di Roma. 
This work was partially supported by: i) PRIN 2022 MUR 20228MZFAA- AIDA (CUP
C53D23003620008, CUP H53D23003480006),  ii) PNRR MUR project PE0000013-FAIR, iii) PNRR – DM 118/2023, iv) Kempe Foundation under grant no. JCSMK24-0094, v) \textit{Strategic University Projects} ``IDEA: AI-powered Digital Twin for next-generation lung cancEr cAre'' financed by \textit{Università Campus Bio- Medico di Roma} 2023 (GEN0469).
Resources are provided by the \textit{National Academic Infrastructure for Supercomputing in Sweden} (NAISS) and the \textit{Swedish National Infrastructure for Computing} (SNIC) at Alvis @ C3SE, partially funded by the Swedish Research Council through grant agreements no. 2022-06725 and no. 2018-05973.
The funders had no role in study design, data collection and analysis, decision to publish, or preparation of the manuscript.
 
\section*{Author Contributions}
F.R., C.M.C., V.G., and P.S. conceptualized the study, developed the methodology and software, and performed validation.
F.R., V.G., and P.S. conducted the formal analysis.
F.R., C.M.C., C.T., B.B.Z., B.V., C.M., A.B., E.F., S.R., V.G., and P.S. contributed to investigation.
F.R., C.M.C., V.G., B.B.Z., and P.S. provided resources.
F.R., C.M.C., C.T., C.G., F.M., M.L., E.I., B.B.Z., M.F., A.C., G.P., B.V., and S.R. curated the data.
F.R., S.R., V.G., and P.S. wrote the original draft and performed review and editing.
C.M.C., C.T., F.M., M.L., A.B., C.M., S.R., V.G., and P.S. contributed to visualization.
S.R. and P.S. acquired funding.
S.R., V.G., and P.S. supervised the work.
All authors reviewed and approved the final manuscript.
 
\section*{Competing Interests}
The authors declare no competing financial or non-financial interests.
 
%% --- END DECLARATIONS BLOCK ---

\bibliography{sn-bibliography}% common bib file
\section*{Figure Legends}

\textbf{Figure 1.} \textit{Overview of modality availability and feature-level sparsity within the cohort}.
Green indicates the proportion of available data, whereas red denotes missingness.
Panel (a) reports the availability of each modality (WSI, CT, and tabular data) across patients.
Panel (b) illustrates feature-wise missingness within the tabular modality, highlighting the proportion of absent values for each clinical variable.

\noindent
\textbf{Figure 2.} \textit{Schematic overview of the proposed Multimodal Survival Framework}.
The framework integrates three clinical data streams: Pathology (WSI), Radiology (CT), and Tabular data. 
\textbf{Step-1} extracts fixed embeddings via domain-specific foundation models. 
\textbf{Step-2} processes each modality through a dedicated \texttt{NAIM+ODST} encoder with built-in missing modality handling. 
\textbf{Step-3} implements \texttt{ConcatODST}, where unimodal latent representations are concatenated and passed through a multimodal module to predict the patient-specific hazard $y$.

\noindent
\textbf{Figure 3.} \textit{Prognostic separability in foundation model embedding spaces}.
Two-dimensional UMAP projections of foundation model embeddings for WSI (a), CT (b), and tabular data (c).
Each plot shows the distribution of survivors (green) and non-survivors (red), with kernel density contours and class centroids.
Reported p-values quantify group separation in the embedding space.

\noindent
\textbf{Figure 4.} \textit{Robustness to modality dropout under increasing missingness}.
Performance of multimodal \textbf{intermediate fusion} models as a function of progressive modality removal.
Each panel reports the mean C-index $\pm$ standard error over 5 folds.
Each line represents the performance when a specific modality is removed at test time, with the x-axis denoting the fraction of test samples for which that modality is missing.
Color coding identifies the modality being removed: green corresponds to Tabular, red to CT, and blue to WSI.
Solid lines denote CT-FM encoders, while dashed lines denote Merlin encoders (panels a, b, d).
Panels correspond to different modality combinations: (a) WSI+CT+Tabular, (b) WSI+CT, (c) WSI+Tabular, and (d) CT+Tabular.

\noindent
\textbf{Figure 5.} \textit{Risk-stratified survival analysis across multimodal configurations}.
Stratified Kaplan–Meier survival curves for intermediate fusion models (\texttt{ConcatODST}), with follow-up time (days) on the x-axis and survival probability on the y-axis.
Patients are stratified into \textit{Low-Risk} and \textit{High-Risk} groups according to the predicted 5-year outcome score produced by each model.
The blue curve represents the \textit{Low-Risk} group, while the red curve corresponds to the \textit{High-Risk} group.
Panels correspond to different modality combinations: (a) CT+Tabular, (b) WSI+CT, (c) WSI+CT+Tabular (trimodal), and (d) WSI+Tabular.

\noindent
\textbf{Figure 6.} \textit{Progression-based survival stratification using the trimodal intermediate fusion model.}
Kaplan–Meier curves for progression-free survival and distant metastasis.
Patients are first stratified into \textit{Low-Risk} and \textit{High-Risk} groups using a threshold derived from validation folds, and this threshold is subsequently applied to the test set.
Within each risk group, patients are further separated according to event status (event vs no-event) for the corresponding endpoint.
This stratification highlights differences in survival trajectories conditioned on both predicted risk and observed progression patterns.
Panel (a) reports progression-free survival (PFS), while panel (b) corresponds to distant metastasis.

\end{document}

% ============================================================
% SUPPLEMENTARY INFORMATION
% ============================================================
\setcounter{section}{0}
\renewcommand{\thesection}{S.\arabic{section}}
\renewcommand{\thesubsection}{S.\arabic{section}.\arabic{subsection}}
\renewcommand{\thetable}{S\arabic{table}}
\renewcommand{\thefigure}{S\arabic{figure}}
\setcounter{table}{0}
\setcounter{figure}{0}

\section*{Supplementary Materials}

\section{Ablation Analysis on Multimodal Module Design}\label{supp:ablation_mm}
% MULTIMODAL TABLE
\begin{table}[ht]
  \centering
  \renewcommand{\arraystretch}{0.85}
  
  \label{tab:ctfm_architectures}
  \begin{tabular}{ll@{\hskip 4pt}l@{\hskip 6pt}cccc}
    \toprule
    \textbf{Layer 1} & \textbf{Layer 2} & \textbf{MS} & \textbf{CT+Tab} & \textbf{WSI+CT} & \textbf{WSI+CT+Tab} & \textbf{WSI+Tab} \\
    \midrule
    \multirow{2}{*}{\texttt{FC}} & \multirow{2}{*}{\texttt{FC}} & Un-frozen & 68.52 {\tiny $\pm$ 5.41} & 72.31 {\tiny $\pm$ 4.76} & 73.49 {\tiny $\pm$ 4.55} & 69.39 {\tiny $\pm$ 2.90} \\
    & & Frozen & \underline{68.94} {\tiny $\pm$ 5.88} & \textbf{76.12} {\tiny $\pm$ 4.01} & 70.85 {\tiny $\pm$ 3.66} & 67.54 {\tiny $\pm$ 4.33} \\
    \midrule
    \multirow{2}{*}{\texttt{FC}} & \multirow{2}{*}{\texttt{\textsc{ODST}}} & Un-frozen & \textbf{69.90} {\tiny $\pm$ 5.08} & 72.49 {\tiny $\pm$ 3.50} & 73.09 {\tiny $\pm$ 3.79} & \textbf{73.30} {\tiny $\pm$ 3.23} \\
    & & Frozen & \textbf{69.90} {\tiny $\pm$ 4.94} & \underline{73.38} {\tiny $\pm$ 3.67} & \textbf{74.42} {\tiny $\pm$ 4.25} & \underline{70.63} {\tiny $\pm$ 1.41} \\
    \midrule
    \multirow{2}{*}{\texttt{\textsc{ODST}}} & \multirow{2}{*}{\texttt{FC}} & Un-frozen & 69.18 {\tiny $\pm$ 4.79} & \underline{73.38} {\tiny $\pm$ 4.35} & 73.38 {\tiny $\pm$ 3.66} & 67.48 {\tiny $\pm$ 2.79} \\ & & Frozen & 68.78 {\tiny $\pm$ 5.47} & 72.75 {\tiny $\pm$ 4.50} & 74.02 {\tiny $\pm$ 3.61} & 68.00 {\tiny $\pm$ 1.80} \\ \midrule \multirow{2}{*}{\texttt{\textsc{ODST}}} & \multirow{2}{*}{\texttt{\textsc{ODST}}} & Un-frozen & 69.18 {\tiny $\pm$ 4.79} & 72.97 {\tiny $\pm$ 4.11} & 72.92 {\tiny $\pm$ 4.86} & 68.13 {\tiny $\pm$ 3.59} \\ & & Frozen & 69.26 {\tiny $\pm$ 5.32} & 72.56 {\tiny $\pm$ 3.60} & \underline{74.29} {\tiny $\pm$ 3.71} & 66.04 {\tiny $\pm$ 2.98} \\
    \bottomrule
  \end{tabular}
  \caption{C-index (mean $\pm$ SE, 5-fold CV) for all Concat architecture variants using CT-FM embeddings. Columns correspond to modality combinations. \textbf{Bold} indicates best performance, \underline{underline} indicates second best.}
  
\end{table}

The multimodal module $E_m(\cdot)$ introduced in Step~3 consists of two sequential layers, denoted $L_1$ and $L_2$. 
$L_1$ acts as a bottleneck that projects the concatenated unimodal representations $h_m \in \mathbb{R}^{d_1+d_2+d_3}$ into a reduced-dimensional space (1/3 of the input dimensionality), fusing the three modality-specific vectors into a single representation. 
$L_2$ preserves the input-output dimensionality and operates as a non-linear refinement stage before the final fully connected projection to the scalar risk score $y$.
We ablated the architectural composition of $E_m$ by substituting each layer independently with either an FC layer or an ODST block, yielding four configurations: FC+FC, FC+ODST, ODST+FC, and ODST+ODST. 
Additionally, we evaluated the effect of freezing versus un-freezing the pretrained unimodal encoders during multimodal training. 
Each configuration was assessed under both encoder settings across all four modality combinations (\texttt{CT+Tab}, \texttt{WSI+CT}, \texttt{WSI+CT+Tab}, \texttt{WSI+Tab}). 
Table~\ref{tab:ctfm_architectures} reports C-index values (mean $\pm$ SE, 5-fold CV).
No single architecture dominates across all modality combinations. 
FC+FC achieves the highest C-index for \texttt{WSI+CT} (76.12, frozen),
ODST+ODST peaks on the trimodal configuration (75.29, frozen), and FC+ODST leads on \texttt{CT+Tab} (69.90) and \texttt{WSI+Tab} (73.30).
However, FC+ODST is the only configuration that ranks first or second across all four modality combinations.
Regarding encoder optimization, the frozen setting yields the highest or most stable C-index for FC+ODST in three of four combinations, while matching the un-frozen variant on \texttt{CT+Tab} (69.90).
Freezing the pretrained unimodal modules also preserves the representations learned during Step~2 and reduces the risk of overfitting on the limited cohort.
These observations motivated the selection of FC+ODST for the MM with frozen unimodal encoders as the final intermediate fusion strategy, referred as ConcatODST.

\section{ODST Implementation and Integration}\label{supp:odst_details}

The Oblivious Differentiable Sparsemax Tree (ODST) block~\cite{popov2019neural} is used throughout the proposed framework: as the survival head $E_i(\cdot)$ in each unimodal branch (Step~2) and as the refinement layer $L_2$ within the multimodal module $E_m(\cdot)$ (Step~3).

Each ODST block implements a differentiable ensemble of oblivious decision trees, in which all nodes at a given depth share the same splitting feature and threshold.
Feature selection is governed by learnable logits
$\mathbf{f} \in \mathbb{R}^{d_{\text{in}} \times T \times D}$, passed through a \texttt{sparsemax} activation~\cite{popov2019neural} that produces a sparse probability vector over input features, summing to one along the feature axis.
For each selected feature, the split decision is computed as
$(\mathbf{v}_j - \tau) \cdot \exp(-\log \mathcal{T})$, where $\tau$ and $\log \mathcal{T}$ are per-node learnable thresholds and log-temperatures, respectively. 
The resulting logits are mapped to approximate binary routing indicators via \texttt{sparsemoid}, a piecewise-linear surrogate of the sigmoid that yields exact zeros outside a bounded interval, enforcing sparse tree routing.
Leaf responses are aggregated by computing the element-wise product of bin matches across all $D$ depths, then contracting with a learnable response tensor $\mathbf{R} \in \mathbb{R}^{T \times d_{\text{out}} \times 2^D}$.

Both thresholds and temperatures are initialized in a data-aware fashion.
Thresholds are set to random quantiles of the training distribution, sampled from a
$\mathrm{Beta}(\beta, \beta)$ prior with $\beta = 1.0$ (uniform). Temperatures are
calibrated so that the majority of data points fall within the linear region of the
\texttt{sparsemoid} function, controlled by a cutoff parameter set to $1.0$.
In all experiments, the ODST ensemble comprises $T{=}32$ trees with depth $D{=}6$ and per-tree output dimension $d_{\text{out}}{=}1$. All ODST hyperparameters are held constant across
modality configurations and fusion strategies.

\newpage
\section{Impact of FM Extraction Module on CT Unimodal and Multimodal Modeling}\label{supp:ct_fms}

\begin{figure*}[h]
    \centering
    \includegraphics[width=\textwidth]{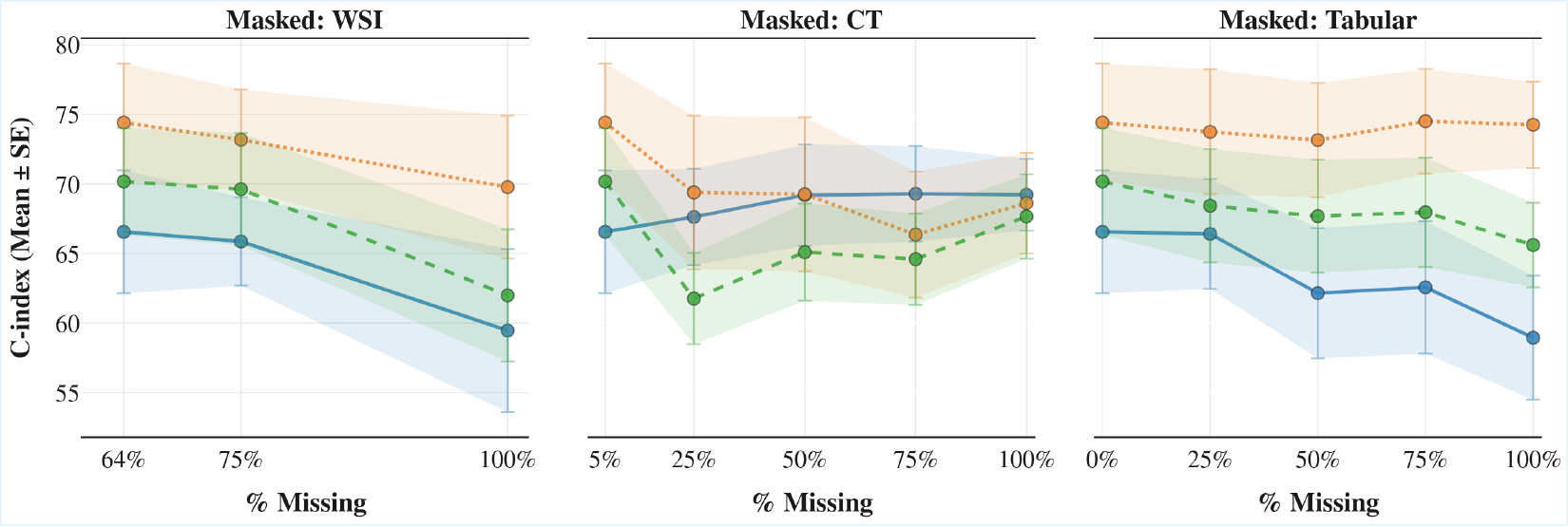}
    \caption{
    }
    \label{fig:trimodal_robustness}
\end{figure*}

The choice of foundation model for CT feature extraction has a direct impact on both unimodal and multimodal survival prediction. 
To quantify this effect, we compared three CT-specific FMs that differ in pretraining strategy and output dimensionality: CT-CLIP~\cite{hamamci2026generalist}, a contrastive vision-language model trained on paired CT volumes and radiology reports yielding a 512-dimensional embedding; Merlin~\cite{MERLIN}, a 3D Vision Transformer pretrained via a CLIP-style objective on large-scale CT, report pairs, producing a 2048-dimensional representation; and CT-FM~\cite{pai2025vision}, a vision transformer pretrained with masked image modeling on over $148,000$ CT volumes, outputting a 512-dimensional class token. 
All three FMs encode the full thoracic volume without tumor-specific region selection, isolating the effect of pretraining strategy from input pre-processing choices.

\begin{table*}[!htbp]
\centering
\small
\renewcommand{\arraystretch}{1}
\setlength{\tabcolsep}{2pt}

\label{tab:unimodal_merged_supp}
\setlength{\tabcolsep}{5pt}
\resizebox{\textwidth}{!}{
\begin{tabular}{llcccc}
\toprule
\textbf{Modality} & \textbf{Model} & \textbf{Type} & \textbf{C-index}  & \textbf{Uno C-index} & \textbf{td-AUC}\\
\midrule
\multirow{7}{*}{\textbf{CT (Merlin)}}
 & \texttt{CPH}       & \texttt{ML}   & 52.84 \tiny{± 2.53} & 53.51 \tiny{± 2.39} & 58.14 \tiny{± 2.34} \\
 & \texttt{RSF}       & \texttt{ML}   & 51.21 \tiny{± 2.74} & 53.33 \tiny{± 1.71} & 54.65 \tiny{± 3.72} \\
 & \texttt{SGB}       & \texttt{ML}   & 52.35 \tiny{± 1.70} & 51.55 \tiny{± 2.37} & 54.39 \tiny{± 3.57} \\
\cmidrule{2-6}
 & \texttt{NN}      & \texttt{DL} & 56.98 \tiny{± 4.44} & 57.39 \tiny{± 4.64} & 63.63 \tiny{± 6.11} \\
 & \texttt{NAIM}     & \texttt{DL} & 53.69 \tiny{± 5.25} & 54.57 \tiny{± 2.06} & 54.23 \tiny{± 3.18} \\
 & \texttt{NAIM+ODST} & \texttt{DL} & 62.20 \tiny{± 4.12} & 63.17 \tiny{± 3.06} & 64.69 \tiny{± 3.87} \\
\midrule
\multirow{7}{*}{\textbf{CT (CT-CLIP)}}
& \texttt{CPH} & \texttt{ML}   & 50.38 \tiny{± 1.91} & 49.90 \tiny{± 1.82} & 49.79 \tiny{± 1.36} \\
& \texttt{RSF}  & \texttt{ML}   & 59.29 \tiny{± 3.12} & 62.22 \tiny{± 3.20} & 57.96 \tiny{± 3.14} \\
& \texttt{SGB}         & \texttt{ML}   & 55.62 \tiny{± 4.61} & 57.98 \tiny{± 5.74} & 57.91 \tiny{± 4.28} \\
\cmidrule{2-6}
& \texttt{NN}        & \texttt{DL} & 57.41 \tiny{± 3.45} & 61.23 \tiny{± 4.82} & 58.12 \tiny{± 3.31} \\
& \texttt{NAIM}      & \texttt{DL} & 55.14 \tiny{± 4.01} & 58.73 \tiny{± 5.42} & 54.86 \tiny{± 3.15} \\
& \texttt{NAIM+ODST} & \texttt{DL} & 56.28 \tiny{± 3.72} & 59.96 \tiny{± 5.17} & 56.39 \tiny{± 2.98} \\
\midrule
\multirow{7}{*}{\textbf{CT (CT-FM)}}
& \texttt{CPH}       & \texttt{ML} & 58.13 \tiny{± 2.51} & 60.78 \tiny{± 2.28} & 56.88 \tiny{± 2.12} \\
& \texttt{RSF}       & \texttt{ML} & 66.62 \tiny{± 6.09} & 71.90 \tiny{± 7.71} & 61.38 \tiny{± 5.15} \\
& \texttt{SGB}       & \texttt{ML} & \textbf{70.22 \tiny{± 5.05}} &  \underline{74.48 \tiny{± 5.76}} & \textbf{66.24 \tiny{± 4.37}} \\
\cmidrule{2-6}
 & \texttt{NN       } & \texttt{DL} & 59.98 \tiny{± 4.24} & 59.39 \tiny{± 3.84} & 58.63 \tiny{± 4.81} \\
 & \texttt{NAIM     } & \texttt{DL} & 63.39 \tiny{± 2.25} & 61.57 \tiny{± 2.16} & 61.23 \tiny{± 3.48} \\
 & \texttt{NAIM+ODST} & \texttt{DL} &  \underline{68.75 \tiny{± 5.29}} & \textbf{74.57 \tiny{± 6.42}} &  \underline{64.42 \tiny{± 4.86}} \\
\midrule

\end{tabular}

}
\caption{Unified unimodal performance across WSI, CT and Tabular modalities. 
Classical \texttt{ML}   models (\texttt{CPH}, \texttt{RSF}, \textbf{SGB}). 
\texttt{DL} architectures (\texttt{NN}, \texttt{NAIM}, \texttt{NAIM+ODST}). 
Mean $\pm$ standard error of the mean (SEM, 5-fold CV).
The best-performing model for each modality is highlighted in \textbf{bold}.}
\end{table*}

Table~\ref{tab:unimodal_merged_supp} reports unimodal survival performance for all six models across the three CT embeddings. 
CT-FM yields the strongest results across both ML and DL model families. 
Among ML baselines, \texttt{SGB} trained on CT-FM features reaches a C-index of $70.22 \pm 5.05$ and Uno C-index of $74.48 \pm 5.76$, compared to $52.35 \pm 1.70$ and $51.55 \pm 2.37$ with Merlin, and $55.62 \pm 4.61$ and $57.98 \pm 5.74$ with CT-CLIP. 
Among DL models, \texttt{NAIM+ODST} with CT-FM achieves a C-index of $68.75 \pm 5.29$, surpassing Merlin ($62.20 \pm 4.12$) and CT-CLIP ($56.28 \pm 3.72$) by $6.55$ and $12.47$ points, respectively. 
The training with CT-CLIP embeddings yelded the minimal discriminative signal. 
\texttt{CPH} model reaches a C-index of $50.38 \pm 1.91$. 
While Merlin occupies an intermediate position; its ML models (C-index $\leq 52.84$) fall below RSF trained on CT-CLIP's embeddings ($59.29$), yet its \texttt{NAIM+ODST} ($62.20$) exceeds the best CT-CLIP configuration, suggesting that Merlin's higher-dimensional representation ($d = 2048$) benefits from the non-linear interactions captured by the \texttt{ODST} layer, while ML models struggle on the larger input space given the limited cohort size.

These differences propagate into multimodal modeling. Figure~\ref{fig:trimodal_robustness} reports the C-index of the trimodal intermediate fusion model (\texttt{ConcatODST}, \texttt{WSI+CT+Tab}) under progressive masking of each modality, with volumetric embeddings sourced from each of the three CT FMs. 
CT-FM (orange) maintains the highest C-index across all three masking conditions and all missingness levels. 
In the masked WSI panel, CT-FM starts at $\approx 75$ and remains above $70$ up to complete WSI removal, while CT-CLIP and Merlin drop below $63$ and $59$ respectively.
In the Masked CT panel, CT-FM declines from $\approx 74$ at natural missingness ($5\%$) to $\approx 65$ at $100\%$, whereas Merlin and CT-CLIP fall to $\approx 60$ and $\approx 63$, respectively. 
In the Masked Tabular panel, CT-FM sustains a C-index above $65$ until $75\%$ missing, while Merlin drops below $60$ at the same threshold. CT-CLIP tracks between the two across all panels.

Two observations follow from this comparison. 
First, masking CT in the trimodal model (Figure~\ref{fig:trimodal_robustness}, middle panel) produces a larger absolute drop for CT-FM (from $\approx 74$ to $\approx 69$) than for Merlin or CT-CLIP, both of which converge to the same endpoint ($\approx 69$) from lower starting values. 
CT-FM does not prevent the performance decline upon CT removal; rather, it raises the baseline from which that decline occurs, indicating that its embeddings encode additional prognostic content that the fusion head exploits when CT is available. 
Second, the Masked Tabular panel reveals a complementary pattern. The gradual removal of the Tabular modality data causes a steep degradation for Merlin (from $\approx 67$ to $\approx 58$) but has minimal effect on the CT-FM configuration (from $\approx 74$ to $\approx 72$). 
This behavior suggests that when the CT backbone provides richer volumetric representations, the fusion model shifts its reliance away from clinical variables, whereas weaker CT embeddings force the model to depend on Tabular features as the primary prognostic driver. 
The overall performance gain from CT-FM therefore stems not from CT alone, but from a redistribution of modality importance within the trimodal model that reduces vulnerability to any single data stream.

\bibliography{sn-bibliography}% common bib file
\section*{Figure Legends}

\textbf{Figure S1.} \textit{Overview of modality availability and feature-level sparsity within the cohort}.
Robustness analysis of the intermediate fusion model ConcatODST, under increasing missingness for each modality.
Each panel shows the C-index (mean $\pm$ SE, 5-fold CV) as a function of the masking rate applied to a single modality while keeping the remaining two at their natural missingness levels (WSI 64\%, CT 5\%, Tabular 0\%).
Three CT foundation models are compared as sources of volumetric embeddings: Merlin , CT-FM, and CT-CLIP.
In the legend, colors denote the corresponding trimodal configurations with the orange line indicating the setting with the CT-FM embeddings, blue for Merlin, and green corresponds to CT-CLIP.